\documentclass[sigplan,nonacm]{acmart}
\AtBeginDocument{%
  }

\usepackage{xspace}
\usepackage{xcolor}
\usepackage{colortbl}
\definecolor{cmpbound}{HTML}{E2E8ED}
\definecolor{membound}{HTML}{F1E3DF}
\usepackage{tabularx}
\usepackage{arydshln}

\newcommand{\sys}{rMuscle\xspace}

\newcommand{\heading}[1]{ \vspace{0.5ex} \noindent \textbf{#1}}

\newenvironment{myitemize}%
  {\begin{list}{\labelitemi}{\itemsep1pt \topsep2pt \parsep0.00in
  \partopsep=0pt \leftmargin1em}}%
  {\end{list}}

\setcopyright{none}
\copyrightyear{2026}
\acmYear{2026}

\usepackage{titlesec}
\titlespacing*{\section}{0pt}{*0.9}{*0.9}
\titlespacing*{\subsection}{0pt}{*0.9}{*0.9}
\titlespacing*{\subsubsection}{0pt}{*0.9}{*0.9}

\AtBeginDocument{%
  \setlength{\abovedisplayskip}{3pt plus 1pt minus 1pt}%
  \setlength{\belowdisplayskip}{3pt plus 1pt minus 1pt}%
  \setlength{\abovedisplayshortskip}{0pt plus 1pt}%
  \setlength{\belowdisplayshortskip}{2pt plus 1pt minus 1pt}%
  \fancypagestyle{standardpagestyle}{%
    \fancyhf{}%
    \fancyfoot[C]{\footnotesize\thepage}%
  }%
  \pagestyle{standardpagestyle}%
}

\makeatletter
\renewcommand\noindentparagraph{\@startsection{paragraph}{4}{\z@}%
  {-.2\baselineskip \@plus -1\p@ \@minus -.2\p@}
  {-3.5\p@}%
  {\ACM@NRadjust{\@parfont}}}
\makeatother

\begin{document}

\title{rMuscle: Robotic Muscle Memory for Efficient Vision-Language-Action Model Inference}

\author{Kaijun Zhou,
  Zhiyang Li,
  Le Chen,
  and Jinyu Gu}
\authornote{Corresponding author: \texttt{gujinyu@sjtu.edu.cn}.}
\affiliation{%
  \vspace{8pt}
  \institution{Institute of Parallel and Distributed Systems, Shanghai Jiao Tong University}
  \country{}
}

\begin{abstract}
  Factory work is a promising early scenario for embodied AI:
assigning repetitive manual jobs to robots has clear economic payoff,
and a structured station keeps the jobs tractable for current policies.
Vision-Language-Action (VLA) models now dominate as the policy paradigm for these robots.
The inference latency of VLA models directly affects robot responsiveness and motion smoothness.
However, existing VLA inference frameworks do not fully exploit the characteristics of embodied workloads or
account for the distinct bottlenecks across different stages of VLA inference.

In this paper, we first characterize embodied workloads and identify substantial task similarity across repeated robot executions.
We further find that such similarity extends beyond observations and action trajectories to internal model states.
Drawing on these observations, we present \textbf{\sys{}}, a real-time VLA inference framework inspired by human muscle memory.
It exploits cross-execution similarity through a dual-phase muscle-memory cache.
The Context Cache reuses visual-token outputs to reduce computation,
while the Action Cache reuses neuron activation patterns to reduce weight accesses.
We keep both the cache memory footprint and access overhead low through online cache recomputation,
sliding-window cache retrieval, and mask sharing across consecutive denoising steps.
\textbf{\sys{}} achieves $1.29$--$1.42\times$ speedup on RTX~4090 and Jetson Thor across LIBERO, RoboTwin,
and physical manipulation tasks,
while maintaining the original success rates on real-world robots.

\end{abstract}

\maketitle

\section{Introduction}
\label{sec:intro}

\begin{figure}[t]
  \centering
  \setlength{\abovecaptionskip}{10pt}
  \setlength{\belowcaptionskip}{0pt}
  \includegraphics[width=\columnwidth]{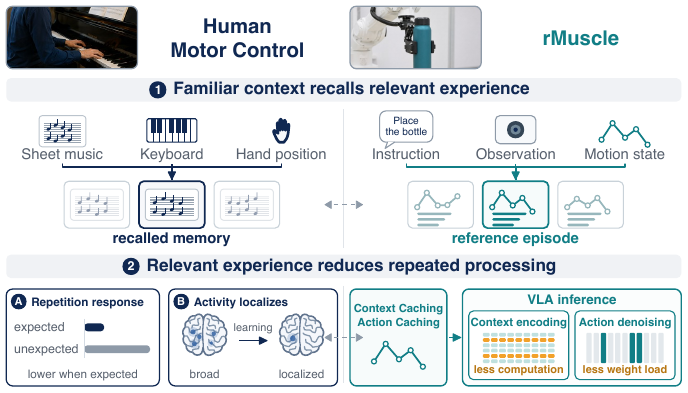}
  \Description{Two stages compared side by side for human motor control and \sys{}.
  Stage one: a pianist's sheet music, keyboard, and hand position recall one learned memory
  from several similar candidates, while a robot's instruction, observation, and motion state
  retrieve one reference episode from several similar candidates. Stage two: two schematics
  show that expected repetitions evoke a lower neural response and that brain activity
  localizes with learning, while \sys{} reuses reference states through its Context and
  Action Caches and recomputes only the parts that changed.}
  \caption{\sys{} operates in a manner analogous to human muscle memory:
  similar contexts trigger relevant prior experience,
  allowing cached results to be retrieved and reused to avoid redundant computation.  }
  \label{fig:cache-analogy}
\end{figure}

Embodied AI extends learned intelligence from digital content to physical interaction,
and factory floors are a promising entry point:
robots that take over repetitive manual tasks offer clear economic value,
and a factory workstation is far more structured than an open-world environment,
with a recurring set of tasks, a finite set of objects, and controlled lighting.
Leading embodied AI companies are therefore bringing robots into factories,
including Figure AI's humanoids on a BMW production line, Tesla's Optimus,
and AgiBot's G2 on a consumer-electronics line~\cite{figurebmw,teslaoptimus,agibotg2}.

Robot execution follows a closed-loop \textbf{observe--infer--execute cycle},
in which sensors capture the current state, a policy predicts an action chunk, and the robot executes it.
Inference latency of the policy therefore limits how frequently the robot can update its actions:
slow inference delays responses to environment changes and can disrupt motion continuity.
The policy typically uses a Vision-Language-Action (VLA) model~\cite{pi0,pi05,groot,xvla,cogact,gr3,go1,smolvla},
which combines multimodal context encoding with diffusion-based generative action prediction
to drive customized intelligent equipment for non-standard automation in factories.

Prior works~\cite{flashrt,vlacpp} develop inference engines to reduce VLA inference latency through kernel fusion, quantization, and static graph execution.
Other studies~\cite{vlaperf2,efficientvla,vla-cache} accelerate VLA inference
by adapting techniques originally developed for VLMs or text-to-image models, such as cross-frame visual-token reuse and denoising-step skipping.
However, these approaches fall short of fully exploiting the acceleration opportunities in VLA inference, primarily due to two limitations.

First, they fail to exploit similarity across embodied tasks.
In industrial manipulation, a robot typically executes a recurring set of tasks at a fixed workstation,
and executions of similar tasks have similar observations and action trajectories.
Such similarity persists across the robot's entire operational workflow,
while existing works focus only on the immediately preceding frame or within a single denoising phase.
Second, existing approaches do not account for the distinct performance bottlenecks across different stages of VLA inference.
In representative models such as $\pi_{0.5}$~\cite{pi05}, multimodal context encoding is predominantly compute-bound,
whereas action denoising is memory-bound, with both stages contributing substantially to end-to-end latency.
Optimizing only one stage results in limited end-to-end speedup.

After an in-depth analysis of VLA workloads and inference behavior,
we find that similarity across task executions extends beyond observations and action trajectories to internal model states,
including FFN inputs and outputs as well as neuron activation patterns.
Inspired by human motor control and muscle memory~\cite{heald2021contextual,summerfield2008repetition,imamizu2000cerebellar} (Figure~\ref{fig:cache-analogy}),
we present \sys{}, a VLA inference framework that exploits this similarity
through a dual-phase cache design tailored to the distinct bottlenecks
of multimodal context encoding and action denoising.
The dual-phase cache consists of a Context Cache and an Action Cache, targeting the two major stages of VLA inference.

The Context Cache accelerates visual context encoding by exploiting similarity in the inputs to latency-critical MLP layers.
Instead of fully recomputing similar inputs, \sys{} reuses their cached outputs and performs computation only for the parts that deviate substantially.
The newly computed outputs are then merged with the cached results to reconstruct the complete layer output,
thereby reducing computation while preserving model accuracy.

The Action Cache targets memory-bound action denoising,
where short sequences make each step dominated by weight loading rather than computation.
Its design builds on two observations: neurons contribute unequally to the result,
and the same denoising step across similar tasks tends to activate similar sets of neurons.
\sys{} thus uses the activation pattern of a similar past execution to predict the important neurons,
loads and computes only their weights,
and approximates the remaining neurons with the outputs of an earlier fully computed step,
substantially reducing memory traffic while retaining the effect of every neuron.

We further introduce three techniques to efficiently manage the dual-phase cache.
First, leveraging the alternating pattern of inference and physical execution in robotic workflows,
\sys{} stores only the input context for caches that are not currently referenced
and reconstructs the complete cache on demand through online recomputation, which is overlapped with action execution.
Second, \sys{} exploits the temporal continuity of robotic workflows.
During the execution of a complete task, the caches that are likely to be referenced in the near future are predictable.
We therefore maintain only a sliding window of relevant caches in GPU memory
and update this window dynamically as the robot executes actions.
Third, \sys{} divides consecutive denoising steps into groups,
with each group sharing an important-neuron mask.
As a result, \sys{} achieves a minimal cache memory footprint and access overhead.

We evaluate \sys{} on $\pi_{0.5}$, GR00T N1.6, and X-VLA across 92 embodied tasks,
spanning 40 LIBERO tasks~\cite{libero}, 50 RoboTwin tasks~\cite{robotwin}, and two
physical-robot tasks: dual-arm pick-and-place on ALOHA~\cite{ALOHA} and single-arm
assembly-line packing with DOBOT and Franka~\cite{dobot,fr3}.
It reaches $1.20$--$1.29\times$ end-to-end speedup on RTX~4090 and $1.23$--$1.42\times$
on Jetson Thor over the state-of-the-art (SOTA) inference engine~\cite{flashrt}.
Our method matches the original success rates across simulation and real-world tasks,
including dual-arm ALOHA manipulation and single-arm assembly-line packing.

This paper makes the following contributions:
\begin{myitemize}
  \item We find that repeated executions of the same task have similar
    FFN inputs and neuron activation patterns, providing opportunities
    for computation reuse.
  \item We design an end-to-end dual-cache inference system for
    SOTA VLA models, with a Context Cache for compute-bound
    VLM prefill and an Action Cache for memory-bound denoising.
  \item We evaluate \sys{} on three models, two simulation benchmarks, and 92 tasks on RTX~4090
    and Jetson Thor, reaching up to $1.29\times$ and $1.42\times$ speedup
    over FlashRT, respectively,
    while matching the average accuracy of the original policies.
\end{myitemize}

\section{Background}
\label{sec:bg}

\subsection{Embodied Robots for Industrial Manipulation}

Factories are a promising entry point for embodied robots,
and leading embodied AI companies are bringing robots onto factory floors.
Figure AI deployed humanoids on a BMW production line,
Tesla is developing Optimus for manufacturing,
and AgiBot reports that its G2 humanoid transports and sorts tablets on a consumer-electronics line~\cite{figurebmw,teslaoptimus,agibotg2}.
A factory-floor case study at Siemens adapts $\pi_{0.5}$ to pack accessory bags into cardboard boxes~\cite{siemenspackaging}.
Amazon's Vulcan Pick picks targeted items from fabric storage pods in a warehouse~\cite{vulcanpick}.

Despite differences in task objectives,
these applications share a common execution pattern:
a robot operates at a fixed workstation and repeatedly performs similar tasks,
with variations mainly in the target objects and their locations.
Such settings also impose stringent requirements
on the responsiveness and reliability of the policy.

\subsection{Embodied Intelligence Serving Loop}

An embodied robot operates in a closed observe--infer--execute loop.
At each inference step, the robot collects visual observations,
a language instruction, and its current joint state,
which consists of the joint angles of the robot's degrees of freedom.
Given these inputs, the policy produces an action chunk containing $S$ future actions,
which the robot executes before the next observation and inference.
This loop repeats until the task is completed, forming an episode.
In our evaluation, completing a task typically requires 4--40 inferences per episode.
If an episode contains $N$ inference-and-execution iterations,
the total number of executed actions required to complete the task lies in $\bigl[(N-1)S+1,\,NS\bigr]$.

\subsection{Visual-Language-Action Models}

VLA models map images, language instructions, and robot state to executable actions. 
Representative models include RT-2, OpenVLA, the $\pi$ family, and GR00T.  
RT-2 and OpenVLA represent actions as text tokens in an autoregressive formulation
that transfers web-scale vision-language pretraining to robot control~\cite{rt2,openvla}.
More recent models generate continuous actions through a diffusion process: 
$\pi_{0.5}$~\cite{pi05} combines a VLM with a denoising action expert for broadly generalizable manipulation, while
GR00T N1.6 pairs a Cosmos-based VLM with a diffusion transformer for
humanoid and bimanual control~\cite{grootn16}.  
X-VLA uses a soft-prompted transformer as a scalable cross-embodiment VLA model~\cite{xvla}.
Those VLA policies are commonly evaluated by rollout success rate on manipulation
benchmarks and real-world tasks that emphasize different capabilities.  

\subsection{Dual-Phase Inference in VLAs}

We focus on the VLA models that couple a vision-language model (VLM) with a diffusion action expert, 
an architecture adopted by the majority of recent VLA models~\cite{pi05,groot,xvla,starvla}.

\begin{figure}[t]
  \centering
  \setlength{\abovecaptionskip}{3pt}
  \setlength{\belowcaptionskip}{0pt}
  \includegraphics[width=\columnwidth]{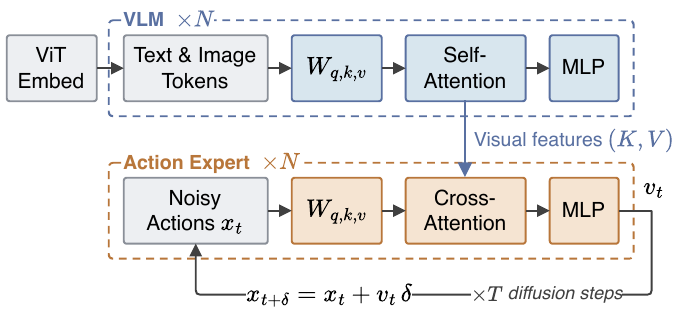}
  \Description{A dual-phase VLA architecture with a VLM stack on the left and a denoising action expert on the right.
  The VLM processes text and image tokens through attention and MLP blocks and passes visual features to the action expert.
  The denoising action expert refines noisy actions through cross-attention and MLP blocks over multiple diffusion steps.}
  \caption{Representative dual-phase VLA architecture $\pi_{0.5}$.
  A VLM processes multimodal inputs and passes visual features to an action expert that iteratively denoises the action chunk.}
  \label{fig:arch}
\end{figure}

As shown in Figure~\ref{fig:arch}, the vision transformer (ViT) embedding stage 
converts the observations and language instruction into input embeddings.
The visual observations typically consist of images captured by the robot's cameras.
Inference then proceeds through two core phases that account for most of the latency.
First, the VLM performs a prefill pass over the ViT embeddings to compute visual features for action generation.
Second, the action expert starts from a noisy action sample and 
iteratively refines it through multiple denoising steps to produce an action chunk.

\subsection{Denoising-Step Skipping and Its Limits in VLAs}

Prior work on accelerating diffusion models for text-to-image and text-to-video generation has used caching
to exploit redundancy at two levels: across denoising steps within a request and across related requests.
DP-Cache and PAB reuse the output of one denoising step as the output of several subsequent steps within a request~\cite{pab,vlaperf2},
while NIRVANA and MoDM maintain references across requests for text-to-image diffusion serving~\cite{nirvana,modm}.
NIRVANA stores intermediate latents from similar prompts and resumes generation from that state,
skipping several denoising steps for new requests~\cite{nirvana}.

However, these methods do not transfer well to embodied model inference,
because both forms of reuse operate at the granularity of denoising steps.
Whole-step reuse relies on consecutive steps producing nearly identical outputs,
as is often the case for text-to-image and text-to-video models that typically run 20--50 steps~\cite{flux,wan}.
VLA models use far fewer steps. For example, GR00T uses only four steps~\cite{grootn16}.
Each step therefore makes a substantial change to the action chunk,
so skipping any step discards refinement that subsequent steps do not recover.

\section{Characterizing Embodied Workloads}
\label{sec:characterizing}

To identify opportunities for accelerating embodied inference, we characterize VLA workloads
from both the execution and model perspectives.
We first locate the performance bottlenecks in VLM prefill and action denoising,
then examine observation and trajectory similarity across executions.
We next examine how this similarity extends to internal model states,
specifically in VLM states and neuron contributions during denoising.

\subsection{Heterogeneous Bottlenecks in VLA Inference}

\begin{figure}[t]
  \centering
  \setlength{\abovecaptionskip}{3pt}
  \setlength{\belowcaptionskip}{0pt}
  \includegraphics[width=\columnwidth]{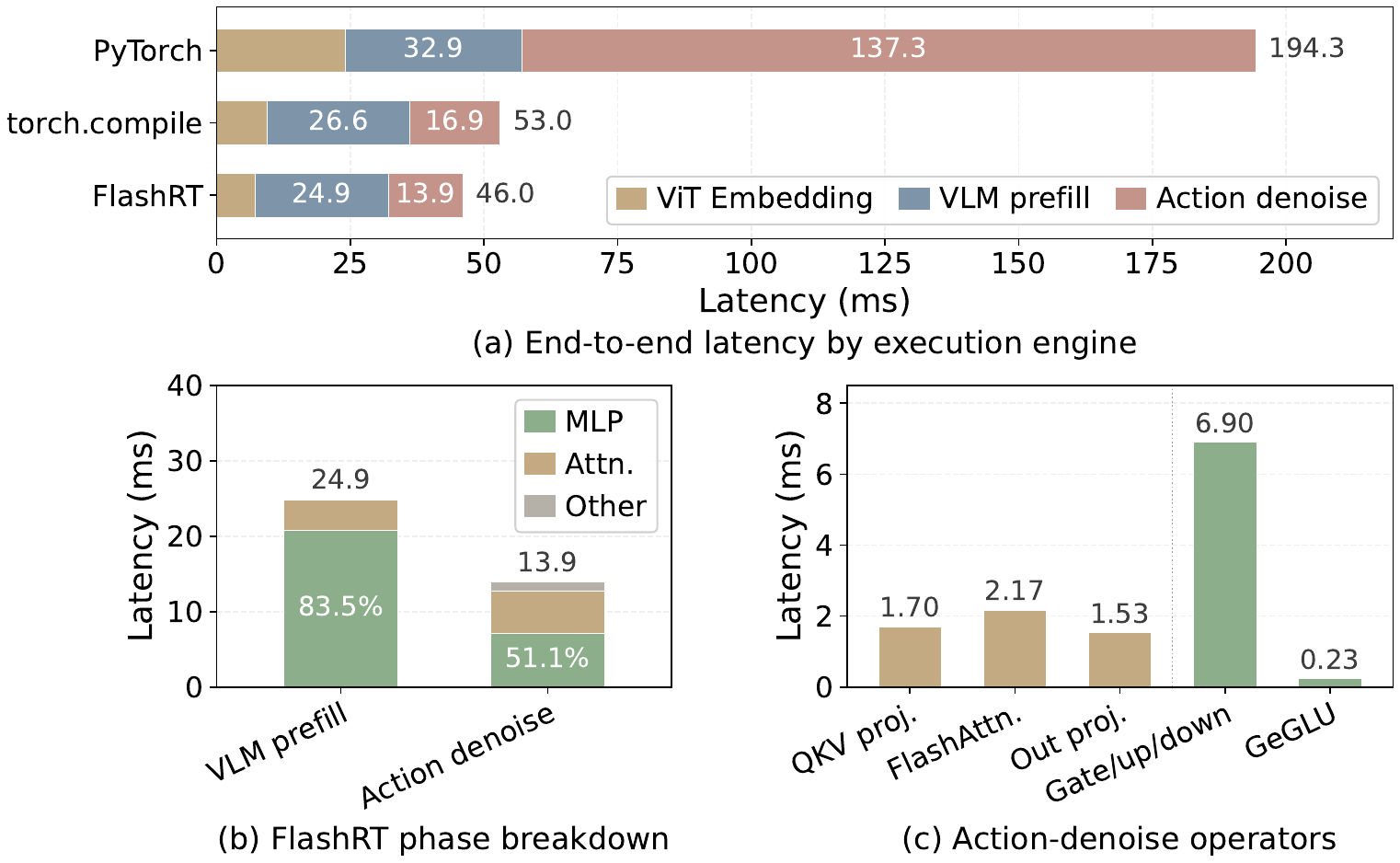}
  \Description{Three panels. Panel (a) shows stacked horizontal bars of end-to-end
  latency for the official PyTorch implementation, torch.compile, and FlashRT, split into
  prefix embedding, VLM prefill, and action denoising. Panel (b) shows stacked vertical bars
  of FlashRT VLM prefill and action denoising latency split into MLP, attention, and other.
  Panel (c) shows per-operator latency aggregated over the action-denoising loop.}
  \caption{Latency breakdown of $\pi_{0.5}$ inference on RTX~4090.
  (a) End-to-end latency with three execution engines.
  (b) FlashRT latency of VLM prefill and action denoising by component.
  (c) Per-operator latency aggregated over the action-denoising loop, colored by component as in (b).}
  \label{fig:latency-breakdown}
\end{figure}

Figure~\ref{fig:latency-breakdown}(a) compares the end-to-end latency across execution engines.
Through kernel fusion and operator optimizations tailored to low-batch workloads, FlashRT reduces the latency from 194.3\,ms with the official PyTorch implementation to 46.0\,ms.
We therefore use FlashRT for the following analysis.

The latency breakdown of FlashRT shows distinct computational profiles across VLM prefill and action denoising phases,
although both spend a substantial fraction of their execution time in feed-forward networks (FFNs).
As shown in Figure~\ref{fig:latency-breakdown}(b),
the MLP accounts for 83.5\% of VLM prefill latency and 51.1\% of action denoising latency.
The operator-level breakdown in Figure~\ref{fig:latency-breakdown}(c) shows that action denoising spends most of its MLP time
in the gate, up, and down projections.
We therefore focus on the MLP in both phases and characterize its bottlenecks next.

\begin{table}[t]
  \setlength{\abovecaptionskip}{5pt}
  \setlength{\belowcaptionskip}{0pt}
  \centering\small
  \caption{Roofline analysis of the $\pi_{0.5}$ MLPs. AI (arithmetic intensity) is in FLOP/B.
  Each device lists theoretical compute / memory latency in $\mu$s derived from BF16 peaks (165\,TFLOP/s, 1000\,GB/s on 4090; 250\,TFLOP/s, 270\,GB/s on Thor).}
  \label{tab:roofline}
  \setlength{\tabcolsep}{2pt}
  \begin{tabular*}{\columnwidth}{@{\extracolsep{\fill}}lrrrcc@{}}
    \toprule
    MLP & GFLOP & MB & AI & RTX 4090 & Jetson Thor \\
    & & & & $T_{\mathrm{comp}}$/$T_{\mathrm{mem}}$ & $T_{\mathrm{comp}}$/$T_{\mathrm{mem}}$ \\
    \midrule
    VLM prefill    & 57.04 & 64 & 891 & \cellcolor{cmpbound}\textbf{345.3} / 64 & 228.2 / 237 \\
    Action denoise & 0.27  & 8  & 34  & \cellcolor{membound}1.6 / \textbf{8.0} & \cellcolor{membound}1.1 / \textbf{29.6} \\
    \bottomrule
  \end{tabular*}
\end{table}

Roofline analysis in Table~\ref{tab:roofline} shows that the MLP faces different bottlenecks in the two phases.
On RTX~4090, the VLM MLP has substantially higher arithmetic intensity and is compute-bound,
whereas the action-denoising MLP has a much smaller workload and is memory-bound~\cite{vlaperf,vlaperf2}.
On Jetson~Thor, the VLM MLP lies near the ridge point: it is theoretically memory-bound,
but the actual bottleneck varies depending on measured hardware performance and implementation details.


\subsection{Trajectory Similarity Across Repeated Tasks}
\label{sec:trajectory-similarity}

In industrial assembly-line settings, robots repeatedly perform the same
task using a limited set of objects in a relatively stable workspace.
However, variations in lighting, tabletop texture, and object pose can
change both observations and required motions.
We therefore examine whether action trajectories remain similar across
executions despite these variations.

\noindentparagraph{Trajectory Similarity.}

\begin{figure}[t]
  \centering
  \setlength{\abovecaptionskip}{3pt}
  \setlength{\belowcaptionskip}{0pt}
  \includegraphics[width=\columnwidth]{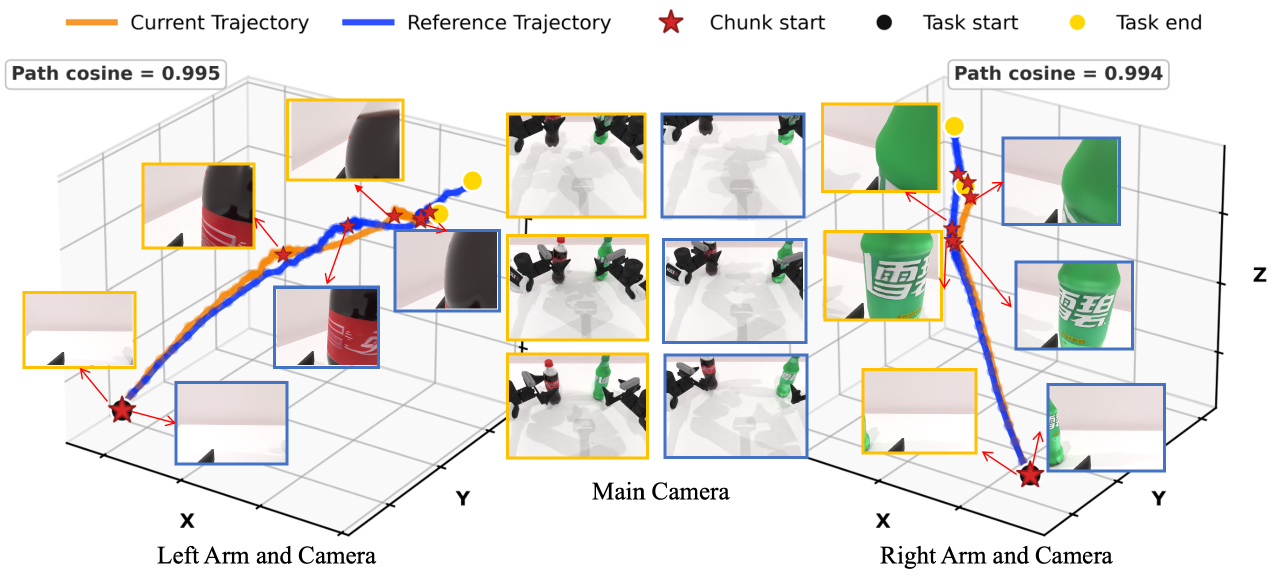}
  \caption{A dual-arm robot moves two soda bottles, one Coke and one Sprite,
  to the same positions above the tabletop.}
  \label{fig:task-repetition}
\end{figure}

An action trajectory is the VLA's observable output: a temporally ordered sequence of robot actions 
that traces a continuous path in 3D space.
To test whether useful repetition survives environmental variation, 
we model these factors in simulation.  
In the representative task shown in Figure~\ref{fig:task-repetition}, a dual-arm robot lifts two beverage bottles
and moves them to designated positions above the tabletop.  
We vary the initial bottle positions and lighting conditions across executions, then
compare a current execution with a reference execution of the same task.

Both the current and reference runs finish in four inference steps and 
reach their target positions successfully and stably.
Although the initial bottle positions and lighting differ, 
the end-effector trajectories and the camera views of both arms remain closely aligned, 
with path cosine similarities of 0.995 for the left arm and 0.994 for the right arm.

\subsection{VLM-State Similarity and Visual-Token Coverage}
\label{sec:characterize-patch}

Comparing MLP inputs and outputs from the current and reference executions at every VLM layer,
we find cosine similarities as high as 98.5\% for inputs and 81.0\% for outputs.
This similarity motivates examining how the remaining output changes are distributed across visual tokens.
We further investigate whether MLP input differences can identify the tokens that account for most of these changes.

\noindentparagraph{Visual-Token Coverage.}

\begin{figure}[t]
  \centering
  \includegraphics[width=\columnwidth]{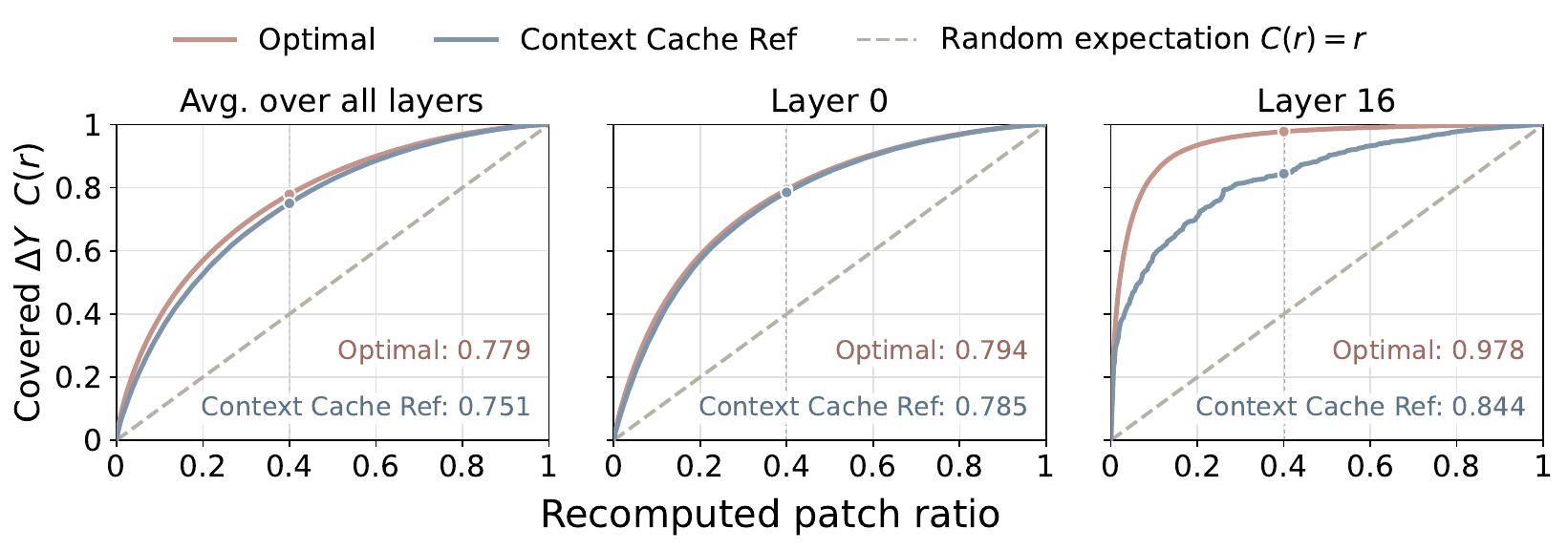}
  \Description{Output-change coverage versus the fraction of selected
  visual patches, averaged across VLM layers and shown for layers 0 and 16.
  Input-change ranking is compared with output-change optimal and random
  selection.}
  \caption{Coverage of $\pi_{0.5}$ VLM FFN output changes under input-based
  patch selection and optimal selection by actual output changes.}
  \label{fig:important-patch}
\end{figure}

At each layer, we rank visual tokens by RMS FFN input change relative to
the current input magnitude and select the top fraction $\rho_{\mathrm{p}}$.
Coverage is the selected tokens' share of the total squared $L_2$ change
in FFN outputs between current and reference executions.
Figure~\ref{fig:important-patch} shows that selecting $\rho_{\mathrm{p}}=0.4$ of the visual
tokens captures approximately 80\% of the output change.

\begin{figure}[t]
  \centering
  \includegraphics[width=\columnwidth]{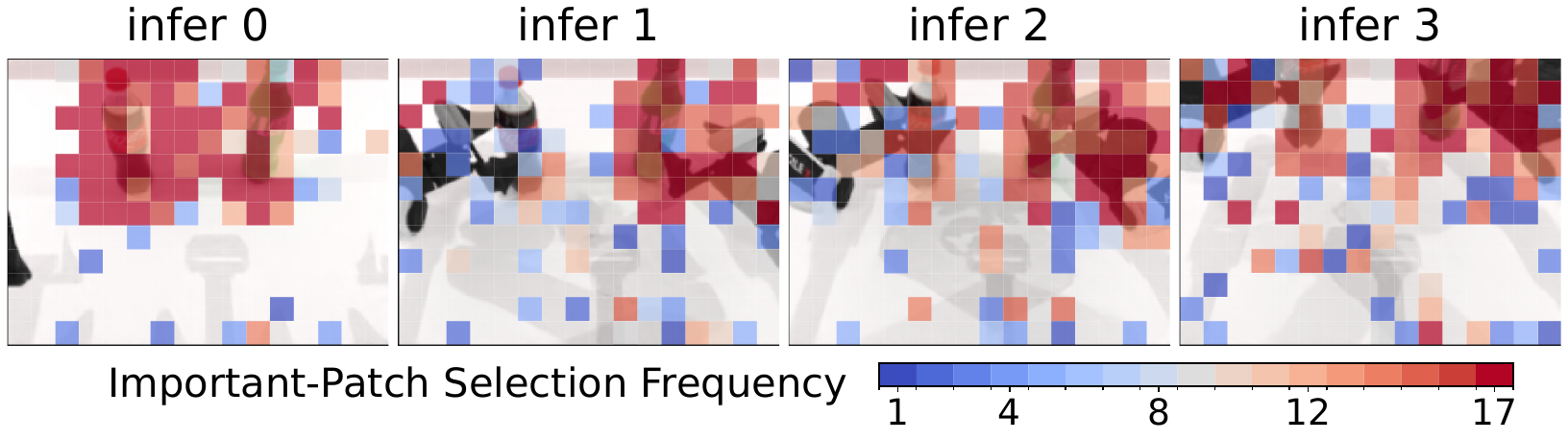}
  \Description{Four main-camera views show the frequency with which
  each visual patch is selected across 17 VLM FFN layers. Selected
  regions move with the bottles, grippers, and arms across action chunks.}
  \caption{Spatial distribution of visual patches selected for
  recomputation across four action chunks.}
  \label{fig:patch_selection}
\end{figure}

Figure~\ref{fig:patch_selection} shows the selected regions for the
bottle-moving task. Important patches concentrate around the bottles,
grippers, and arms, with a smaller number on background regions that
encode scene context. Their locations shift across action chunks as
manipulation progresses, motivating selection for each policy call
rather than a fixed spatial mask.

\subsection{Denoising-Neuron Contribution and Overlap}

The similar trajectories of current and reference executions motivate
examining whether their action-denoising processes also exhibit similar
internal activations. Following our analysis of VLM states, we first
compare the inputs and outputs of action denoising.
Although the initial noise is identical under the same random seed,
the generated action chunks exhibit only 69\% similarity.
This difference motivates exploring finer-grained reuse within denoising
instead of directly reusing reference actions.
We therefore examine how neuron contributions are distributed within
each denoising FFN and whether these contribution patterns are similar
between the current and reference executions.

\noindentparagraph{Neuron Contribution and Coverage.}

\begin{figure}[t]
  \centering
  \setlength{\abovecaptionskip}{3pt}
  \setlength{\belowcaptionskip}{0pt}
  \includegraphics[width=\columnwidth]{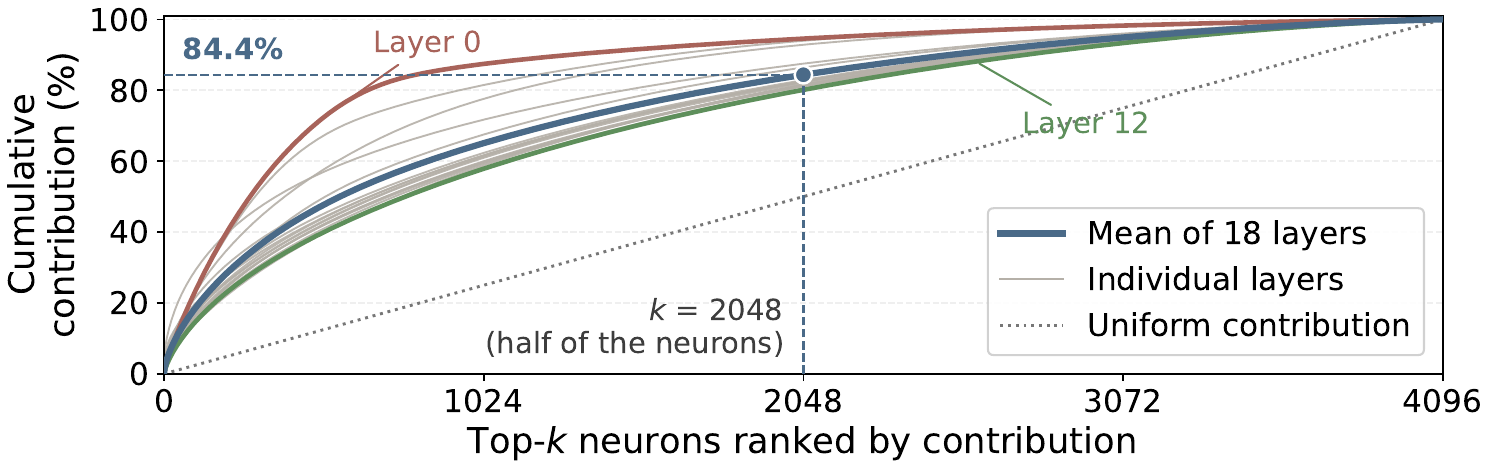}
  \caption{Cumulative contribution-score coverage of denoising MLP neurons in $\pi_{0.5}$.}
  \label{fig:neuron_contribution}
\end{figure}

For each denoising step $t$ and layer $l$, we quantify each neuron's
contribution to the FFN output using the following score~\cite{wanda}:
\begin{equation}
\label{eq:neuron-contrib}
c_{t,l,n}
  = \Bigl(\sum_{r=1}^{S}\lvert h_{t,l,r,n}\rvert\Bigr)
    \lVert W_{l,n}\rVert_2.
\end{equation}
Here, $h_{t,l,r,n}$ is the post-GeGLU activation of neuron $n$ for action $r$, and
$W_{l,n}$ is its down-projection weight.
Within each $(t,l)$, we rank all neurons by $c$ in descending order.
We call the highest-contribution neurons important neurons, and write
$\rho_{\mathrm{n}}$ for the fraction of hidden neurons retained at each layer.
Coverage is the share of the total contribution score captured by the top $\rho_{\mathrm{n}}$ fraction of neurons.
Each layer curve in Figure~\ref{fig:neuron_contribution} averages this
coverage across 100 inferences and denoising steps.

As shown in Figure~\ref{fig:neuron_contribution}, retaining $\rho_{\mathrm{n}}=0.5$
of the neurons captures 84.4\% of the total contribution score.

\noindentparagraph{Neuron Overlap Across Executions.}
\label{sec:neuron-overlap}

We then compare these important-neuron sets across executions.
For repeated executions of the same task, the overlap averages 86\%
(ranging from 82\% to 91\%), whereas for executions of different tasks
it drops to 71\% (70\%--72\%). The gap between the two settings is
larger than the within-setting variation, indicating that important
neurons are task-dependent rather than fixed and motivating the use of a
same-task reference to predict important neurons.


\section{Design}
\label{sec:design}

\noindentparagraph{Overview.}
\sys{} combines two complementary mechanisms: dual-phase caching
for selective recomputation and cache management for low-overhead
access and storage.

\begin{figure}[t]
  \centering
  \includegraphics[width=\columnwidth]{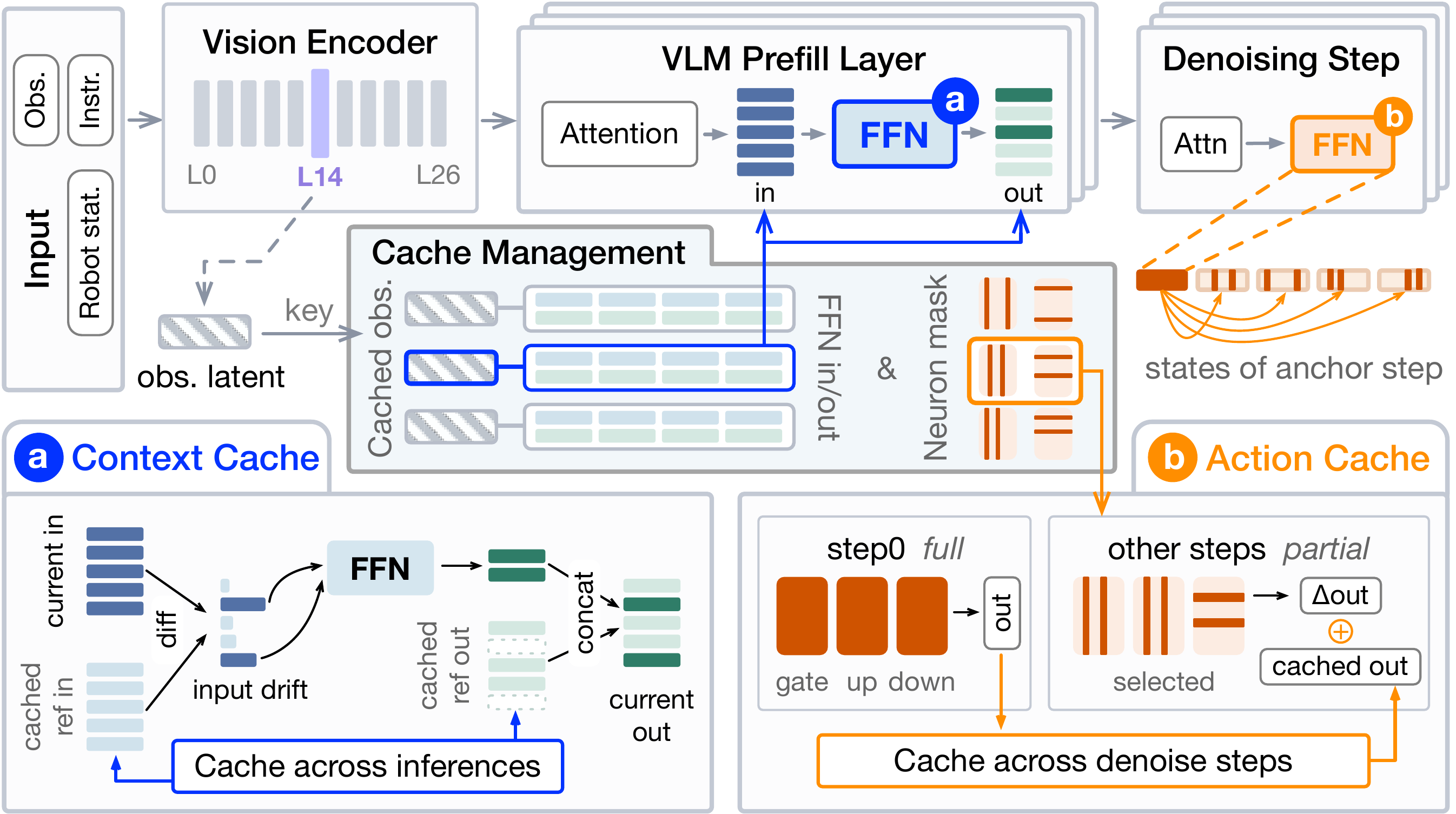}
  \Description{Dual-phase cache architecture. Intermediate vision features
  retrieve a reference for VLM FFN input and output reuse. During action
  denoising, selected neuron updates are added to outputs from dense
  anchors in the current policy call.}
  \caption{Dual-phase cache design. Context Cache reduces visual-token FFN
  computation during VLM prefill. Action Cache reduces weight memory accesses during denoising.}
  \label{fig:overview}
\end{figure}

Figure~\ref{fig:overview} illustrates how the dual-phase cache works.
It uses a Context Cache to accelerate compute-bound VLM prefill,
and an Action Cache for memory-bound action denoising.
The Context Cache reuses FFN outputs for similar visual tokens and
recomputes those whose inputs have changed substantially.
During action denoising, similar executions tend to activate overlapping
sets of important neurons. The Action Cache uses these reference activation
patterns to select which neurons to evaluate and which weights to load.

Cache management supplies the dual-phase cache with reference FFN inputs
and outputs and neuron masks from a shared library.
To reduce memory overhead, the library retains compact reference inputs
and neuron masks, and expanded FFN states are only reconstructed
when a reference is likely to be retrieved.
References are selected using visual context and robot-state history.
A sliding GPU working set is further used to bound memory residency,
combined with a reconstruction budget that limits reconstruction work.

\subsection{Dual-Phase Cache}
\label{sec:design-dual-phase}

Given a reference from the cache library, Context Cache uses its
FFN inputs and outputs for visual-token recomputation. 
Action Cache uses neuron masks to select which important neurons 
to update relative to the current call's dense anchors. 
The following mechanisms specify these computations.

\noindentparagraph{Visual-Token Recomputation.}
\label{sec:design-context}

Our visual-token selective recomputation is inspired by prior work on token-level
and similarity-based feature reuse~\cite{vitreuse,focus,toca}.
We adapt these reuse principles to VLM FFNs across task executions:
the Context Cache recomputes selected visual-token rows and reuses the
remaining outputs from a retrieved reference execution.
Input drift predicts which visual tokens account for most of the
FFN output change, as measured in \S\ref{sec:characterize-patch}.
For visual token $p$ at layer $l$, let $x_{l,p}$ be the current FFN
input after RMSNorm and let $(\bar{x}_{l,p},\bar{y}_{l,p})$ be the
reference input and output. The runtime ranks visual tokens by
\begin{equation}
  d_{l,p}
    = \frac{\operatorname{RMS}(x_{l,p}-\bar{x}_{l,p})}
            {\max\bigl(\operatorname{RMS}(x_{l,p}),\varepsilon\bigr)},
  \; \varepsilon=10^{-6}.
  \label{eq:patch-drift}
\end{equation}
For $P$ visual tokens, the set $\mathcal{P}_l(\rho_{\mathrm{p}})$
contains the $\lceil\rho_{\mathrm{p}}P\rceil$ tokens with the largest
scores. This selection uses only FFN inputs, so it does not require
evaluating the current dense FFN to decide what to reuse.

Selected visual rows are recomputed, and the remaining rows reuse their
reference outputs:
\begin{equation}
  \widehat{y}_{l,p} =
  \begin{cases}
    F_l(x_{l,p}), & p\in\mathcal{P}_l(\rho_{\mathrm{p}}),\\
    \bar{y}_{l,p}, & p\notin\mathcal{P}_l(\rho_{\mathrm{p}}).
  \end{cases}
  \label{eq:patch-update}
\end{equation}
Instruction and robot-state rows are always recomputed.
The selected rows pass through the FFN's gate, up, activation, and down
operations and are then scattered into a full output buffer initialized with
the reference visual outputs.
The profiled ratio $\rho_{\mathrm{p}}$ bounds FFN computation while
allowing selected positions to vary by layer and policy call.
Attention remains dense over the complete sequence.

\noindentparagraph{Group-Shared Neuron Masks.}

Identifying the exact important neurons for the current step through
dense execution would defeat the reduction in weight traffic.
A task-independent mask would instead ignore the task-dependent
activation patterns observed in \S\ref{sec:neuron-overlap}.
\sys{} therefore predicts important neurons from a related reference
and stores their indices with that reference's inputs.

\begin{figure}[t]
  \centering
  \setlength{\abovecaptionskip}{3pt}
  \setlength{\belowcaptionskip}{0pt}
  \includegraphics[width=\columnwidth]{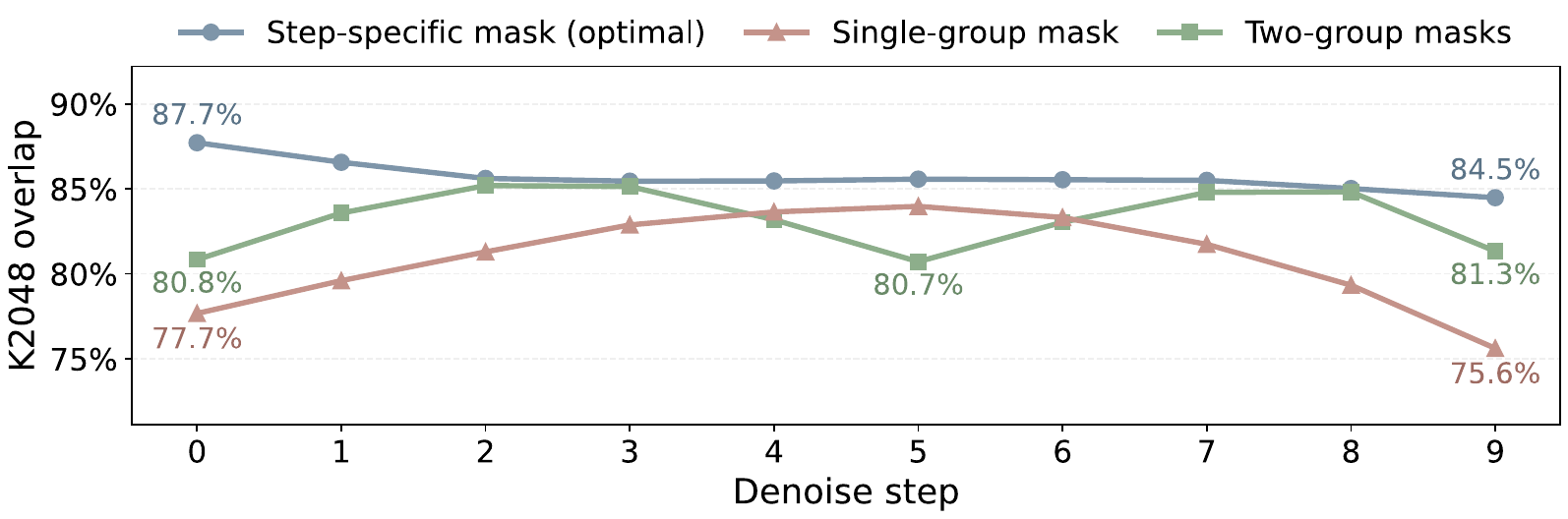}
  \Description{Important-neuron coverage across denoising steps for
  step-specific reference masks and masks shared within one or two groups.}
  \caption{Important-neuron coverage with step-specific and group-shared
  masks, using top-$k=2048$ neurons.}
  \label{fig:neuron-overlap}
\end{figure}

A separate mask for every denoising step requires repeated weight gathering. 
Consecutive denoising steps can instead form a group that shares one mask,
trading some step-specific coverage for less gather traffic.
Figure~\ref{fig:neuron-overlap} compares these choices for $\pi_{0.5}$.
Two group-specific masks track step-specific important neurons more closely than
one mask shared across all ten steps. The first and middle steps run
densely, so mask coverage does not limit computation at those anchors.
Offline mask-construction is described in \S\ref{sec:details}.

\noindentparagraph{Anchors with Mask Recomputation.}
\label{sec:design-overlap}

Each group begins with a dense anchor that establishes the current policy call's activations.
Let $a_g$ be the anchor step of group $g$. At each action-expert layer $l$,
the anchor executes the dense FFN, producing hidden activations $h_{a_g,l}$
and output $y_{a_g,l}$. The Action Cache retains the output and the
activations selected by the group's mask.
In the ten-step $\pi_{0.5}$ configuration, step 0 anchors
for steps 1--4, and step 5 anchors for steps 6--9.

At a non-anchor step $t$, let $M=\mathcal{M}_l^{(g)}$ be the retrieved
reference mask. We compute only the selected neurons' hidden activations
$h_{t,l}^{M}$ and update the output using their down-projection weights
$W_{l,M}^{\mathrm{down}}$:
\begin{equation}
  \widehat{y}_{t,l}
    = y_{a_g,l}
       +\left(h_{t,l}^{M}-h_{a_g,l}^{M}\right)
        W_{l,M}^{\mathrm{down}}.
  \label{eq:local-update}
\end{equation}
Subtracting the selected anchor contribution before adding its current
value avoids double counting and preserves the anchor contribution
of neurons outside $M$.
The output follows the normal adaptive gate and residual addition,
and attention remains dense.
The reference determines which neurons to update, while all reused
anchor values come from the current policy call.
Selected weights are prepared in shared compact buffers and reused across
each group's steps, with gathering overlapped with model execution (\S\ref{sec:details-overlap}).

\subsection{Cache Management}
\label{sec:design-management}

Cache management supplies the dual-phase cache with relevant reference
data while bounding GPU residency and reconstruction work.
It combines asynchronous online reconstruction, context-aware retrieval, and
sliding-window prefetching guided by temporal locality.

\begin{figure}[t]
  \centering
  \setlength{\abovecaptionskip}{3pt}
  \setlength{\belowcaptionskip}{0pt}
  \includegraphics[width=\columnwidth]{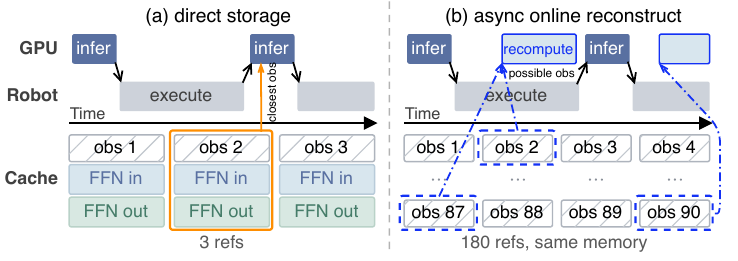}
  \Description{Direct storage retains expanded FFN states for every
  reference. Asynchronous reconstruction stores compact reference inputs
  and prepares selected FFN states while the robot executes actions.}
  \caption{Direct storage versus asynchronous online reconstruction.
  Robot execution provides time to expand compact reference inputs
  before the next policy call.}
  \label{fig:online-compute}
\end{figure}

\noindentparagraph{Asynchronous Online Reconstruction.}

Expanded VLM states are too large to store for every reference across
all tasks. For three-view $\pi_{0.5}$, each of the 17 FFN layers retains
a pair of $[768,2048]$ BF16 input and output tensors, approximately
107\,MB per policy call. An episode of 40 policy calls therefore requires about
4.28\,GB of intermediate states. Storing only compact inputs avoids
this expansion, but reconstructing the states synchronously would put
dense VLM computation back on the inference critical path.

\sys{} instead reconstructs references during physical action execution,
as shown in Figure~\ref{fig:online-compute}.
The repository retains reference model inputs and offline neuron masks.
During an execution window, the GPU fetches inputs for upcoming candidate
positions and runs VLM prefill densely to prepare their FFN states.
Since neuron masks are computed offline, reconstruction does not rerun
action denoising.
Only the active working set holds expanded FFN states on the GPU.
This organization reduces persistent host and disk storage, but does
not reduce the GPU footprint of an individual reconstructed entry.
The candidate positions are determined by the prefetching policy described below.


\noindentparagraph{Retrieving a Relevant Reference.}

The instruction first restricts retrieval to a compatible task pack.
Visual matching uses the policy's intermediate ViT features.
For three-view $\pi_{0.5}$, the query is a BF16 tensor of shape
$[768,1152]$ from the 15th of 27 vision layers.
GPU search ranks reference positions by cosine
similarity between flattened current and reference representations,
retaining the top $T$ candidates. It then compares their aligned
robot-state histories of length $L$, represented as $[L,14]$ tensors,
and selects the candidate with the smallest motion distance to distinguish
motions through similar scenes in different directions.
\S\ref{sec:details} specifies the motion-distance calculation and fallback policy.

Search is limited to references whose FFN states are already prepared on
the GPU, so retrieval requires no transfer of expanded states on the
critical path. The intermediate ViT query allows retrieval to overlap
with the remaining vision encoding. \S\ref{sec:details-overlap}
describes the stream schedule.

\noindentparagraph{Temporal Locality.}

\begin{figure}[t]
  \centering
  \setlength{\abovecaptionskip}{3pt}
  \setlength{\belowcaptionskip}{0pt}
  \includegraphics[width=\columnwidth]{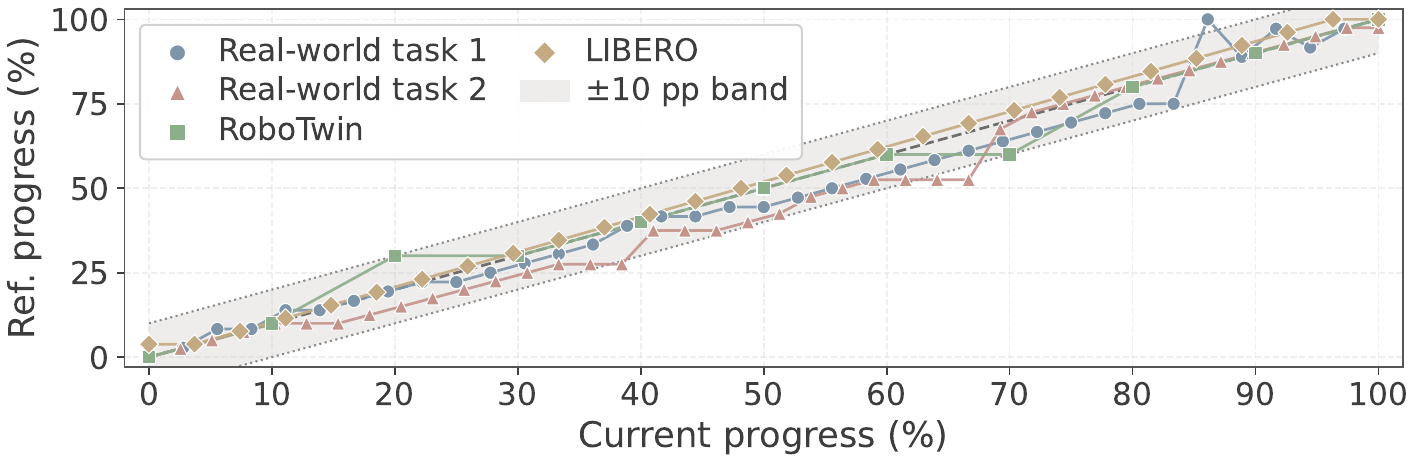}
  \Description{Normalized reference progress versus normalized current
  episode progress. Matches concentrate near the diagonal, with 96.6\%
  within approximately ten percentage points of aligned progress.}
  \caption{Temporal locality of retrieved references. Matches track
  current progress along the diagonal.}
  \label{fig:timewindow}
\end{figure}

Figure~\ref{fig:timewindow} compares reference and current
episode progress. Matches concentrate along $y=x$:
as the current episode proceeds, observations change and retrieval
advances to later references.
In the measured traces, 96.6\% of matches remain within approximately
ten percentage points of aligned progress.
This locality suggests that a small window of reference positions can
cover the references likely to be needed next.

\noindentparagraph{Sliding Window Prefetching.}

\begin{figure}[t]
  \centering
  \setlength{\abovecaptionskip}{3pt}
  \setlength{\belowcaptionskip}{0pt}
  \includegraphics[width=\columnwidth]{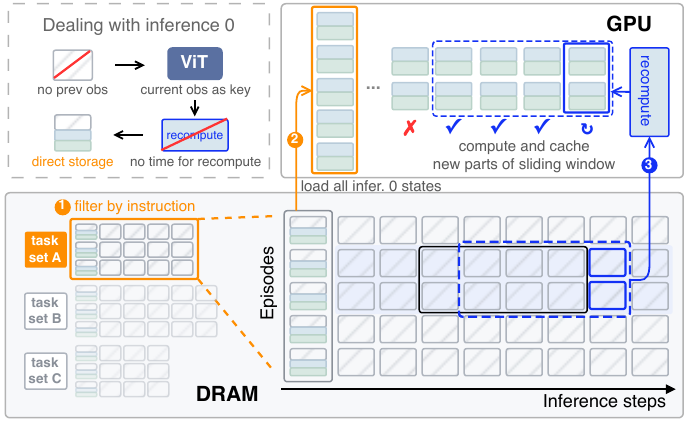}
  \Description{Instruction filtering selects a task pack.
  Initial-position states support the first policy call. A two-dimensional
  window then retains selected reference episodes and nearby positions,
  preparing only new positions as execution advances.}
  \caption{Mechanism of the sliding GPU working set.}
  \label{fig:sliding-window}
\end{figure}

We use this locality to maintain a sliding GPU working set
across candidate episodes and nearby inference positions
(Figure~\ref{fig:sliding-window}).

The first policy call has neither a preceding execution window nor a
previous match. Before an episode starts, 
\sys{} therefore loads the initial-position FFN states of the current task's reference episodes. 
With the robot at its known initial pose, inference 0 selects the top $T$ episodes by visual similarity. 
Only their upcoming positions are then reconstructed during the first action chunk.

The initial visual matches select candidate episodes, and subsequent
matched positions guide the window's progress.
In the representative configuration, the episode-axis
width is $T=4$. For each candidate episode, the window retains the
current matched position and the next $L=4$ positions, yielding
$T(L+1)=20$ entries in total. The four-position lookahead corresponds
to 10\% of the maximum observed episode length of about 40 policy calls.
The robot-state history length is also $L=4$.
\sys{} retains entries shared by successive windows, releases entries
that leave the window, and reconstructs only newly entering positions.
Thus, the persistent repository can grow without making every
reference's intermediate states GPU-resident.

\sys{} also handles exceptional cases during sliding-window execution.
If no suitable reference is available, it invokes the fallback policy.
If the same reference is retrieved more often than the task's maximum number of policy calls,
\sys{} identifies the robot as stalled, resets the episode, and marks the episode as a failure.


\noindentparagraph{Cache Entry Resident Count Budget.}

The cache-entry reconstruction count is bounded by candidate availability, free GPU memory, and
the time before the next policy call. With the parameters in
Table~\ref{tab:recompute-budget}, the number of new entries is
\begin{equation}
  N_{\mathrm{prep}} = \min\!\left\{
    N_{\mathrm{cand}},
    \left\lfloor\frac{M_{\mathrm{free}}}{M_{\mathrm{entry}}}\right\rfloor,
    \left\lfloor\frac{T_{\mathrm{avl}}-T_{\mathrm{trans}}}
                           {\bar{T}_{\mathrm{rec}}}\right\rfloor
  \right\}.
  \label{eq:recompute-budget}
\end{equation}
This expression estimates capacity from average
reconstruction time rather than guaranteeing a hard deadline for each entry.
Asynchronous inference may start the next policy call before the current
action chunk finishes~\cite{rtc}, shortening the reconstruction window ($T_{\mathrm{avl}}$).
\sys{} could adjust the number of reference entries prepared to fit the available time.

\begin{table}[t]
  \centering\small
  \setlength{\abovecaptionskip}{5pt}
  \setlength{\belowcaptionskip}{0pt}
  \setlength{\tabcolsep}{3pt}
  \caption{Parameters for the online reconstruction budget.}
  \label{tab:recompute-budget}
  \begin{tabularx}{\columnwidth}{@{}lX@{}}
    \toprule
    \multicolumn{1}{c}{Symbol} & \multicolumn{1}{c}{Description} \\
    \midrule
    $N_{\mathrm{cand}}$ & New candidate positions entering the window. \\
    $M_{\mathrm{free}}$ & Free GPU memory after reserving model and workspaces. \\
    $M_{\mathrm{entry}}$ & Expanded intermediate-state footprint per entry. \\
    $T_{\mathrm{avl}}$ & Available reconstruction interval. \\
    $T_{\mathrm{trans}}$ & Input-transfer time for the proposed batch. \\
    $\bar{T}_{\mathrm{rec}}$ & Average dense reconstruction time per reference. \\
    $N_{\mathrm{prep}}$ & Number of new entries reconstructed. \\
    \bottomrule
  \end{tabularx}
\end{table}

\section{Implementation}
\label{sec:details}

This section describes how \sys{} builds the cache offline
and prepares and retrieves cached references to accelerate online inference.

\noindentparagraph{Profiling and Constructing the Cache.}
\sys{} uses two recomputation ratios: $\rho_{\mathrm{p}}$ for visual tokens and $\rho_{\mathrm{n}}$ for denoising neurons,
both profiled offline for each model.
Cache construction proceeds in two rounds.
First, we sample task variants across object categories and partition the workspace by object pose, e.g., into four quadrants.
We profile 2--8 episodes per region, depending on task difficulty,
to estimate the recomputation ratios and populate the cache.
Second, we cluster observations from these episodes and collect additional episodes for underrepresented clusters to improve scene coverage.
This process yields 200--800 inference calls per task in our evaluation.



\noindentparagraph{Constructing Group Neuron Masks.}
To construct neuron masks, we run the cached inference offline and compute
$c_{t,l,n}$ using Equation~\ref{eq:neuron-contrib}.
For each layer $l$, we select the top $K_l=\lceil\rho_{\mathrm{n}}H_l\rceil$ neurons at
each denoising step. Within each denoising group, we take the union of these candidates,
rank them by summed contribution, and retain the top $K_l$.
The resulting mask is shared across non-anchor steps in the group.


\noindentparagraph{Cache Capacity and Reconstruction Budget.}
\label{sec:details-capacity}
Online reconstruction is bounded by both execution time and GPU memory.
In the three-view $\pi_{0.5}$ configuration, reconstructing one VLM cache entry takes about 25\,ms,
while the available GPU memory on an RTX~4090 can hold over one hundred reconstructed entries.
A 15-action chunk executed at 30\,Hz provides roughly 500\,ms during which reconstruction can overlap with robot execution,
allowing about 20 entries to be prepared before the next inference request.
Since only compact reconstruction inputs are transferred,
at about 1.7\,MB per entry, data movement contributes little additional overhead.
We therefore determine the reconstruction budget from the available execution interval,
memory capacity, and measured transfer time.



\noindentparagraph{Joint-History Retrieval.}
Subsequent calls retrieve candidate positions from the top $T$
episodes selected at inference 0.
These positions are reranked by the RMS joint-angle difference
between query and candidate histories, computed over aligned
history steps and each arm's joints (e.g., six per arm for ALOHA).
This distance distinguishes execution states that may share similar end-effector poses
but differ in joint configurations.
Single-arm tasks use the active arm's distance, while dual-arm tasks use
the larger of the two arm distances.
The closest candidate provides the cache payload.
\textbf{Fallback policy.} If the visual cosine similarity is below 0.8, \sys{} switches to dense execution
for the remainder of the episode and retries cache reuse in the next episode.



\noindentparagraph{Overlapping Retrieval and Weight Gathering.}
\label{sec:details-overlap}
Once intermediate ViT features are available, a side CUDA stream
retrieves a prepared reference while the main stream completes vision
encoding. The side stream then gathers selected neuron weights,
preparing the first denoising group before the second, while the main
stream executes VLM prefill.
CUDA events ensure that retrieval completes before VLM FFN reuse
and that each denoising FFN waits until its required weights are ready.

\section{Evaluation}
\label{sec:eval}

We evaluate \sys{} to answer three major questions:
\begin{myitemize}
  \item How much speedup does it provide, and how much does each optimization contribute? (\S\ref{sec:eval-performance} \& \S\ref{sec:eval-ablation})
  \item Does it preserve task success rates across diverse tasks, and how do cache configurations affect this? (\S\ref{sec:eval-quality} \& \S\ref{sec:eval-ablation})
  \item What additional memory overhead does it incur? (\S\ref{sec:eval-memory})
\end{myitemize}


\subsection{Experimental Setup}
\label{sec:eval-methodology}

\heading{Hardware and Models.}
We use a desktop with an RTX~4090 (24\,GB GPU memory), an Intel Core i7-11700,
and 32\,GB RAM, as well as a Jetson Thor with 128\,GB unified memory.
They run PyTorch~2.6.0/CUDA~12.4 and PyTorch~2.8.0/CUDA~13.0, respectively.
We evaluate $\pi_{0.5}$, GR00T N1.6, and X-VLA~\cite{pi05,grootn16,xvla}
on LIBERO, and $\pi_{0.5}$ on RoboTwin, following checkpoint availability.
All methods use BF16 or FP16 without quantization.
All $\pi_{0.5}$ latencies use three camera inputs, masking the unused
view on two-camera LIBERO.

These three models represent distinct VLA architectures.
$\pi_{0.5}$ is VLM-heavy, so Context Cache reuse has the larger opportunity.
GR00T N1.6 is denoising-heavy, so Action Cache reuse and denoising-step skipping have the larger opportunity.
X-VLA has a lightweight VLM, but its visual features are concatenated with the denoising actions and processed by self-attention.
We therefore apply visual-token sparsity and neuron sparsity together inside the denoising loop.

\heading{Simulation Benchmarks.}
LIBERO~\cite{libero} covers 10 tasks in each of its Spatial, Object,
Goal, and Long suites, with 100 seeded rollouts per task under the standard
initialization protocol.
For RoboTwin~\cite{robotwin}, we use the clean setting across 50 tasks,
with 100 seeded rollouts per task varying object instances and poses.

\begin{figure}[b]
  \centering
  \setlength{\abovecaptionskip}{3pt}
  \setlength{\belowcaptionskip}{0pt}
  \includegraphics[width=\columnwidth]{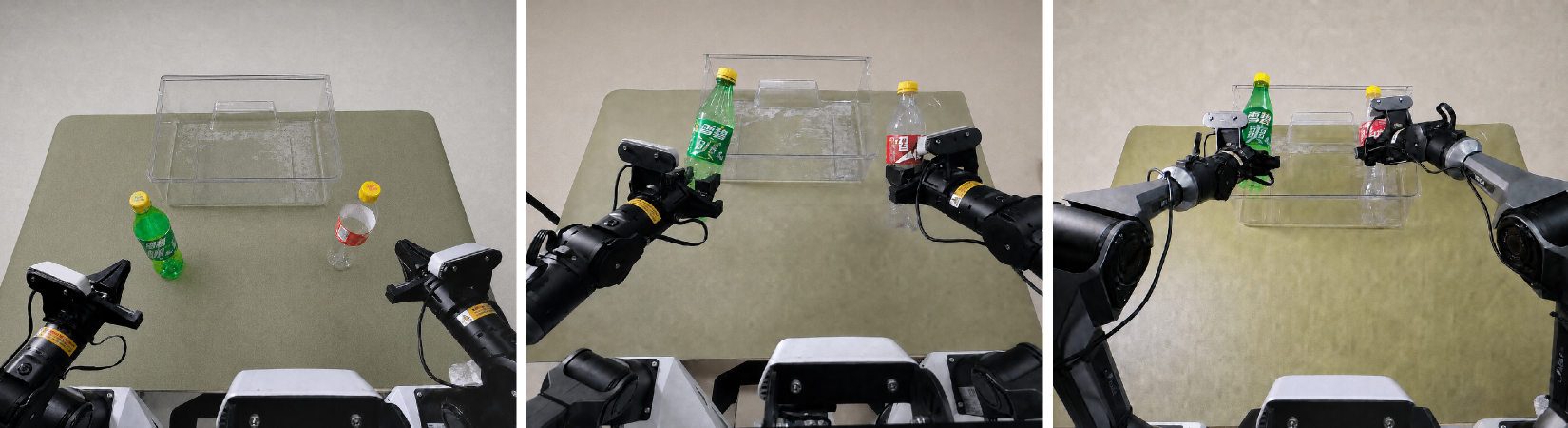}
  \Description{Three views of the cooperative task show the
  initial bottle positions, the dual-arm grasp, and the bottles moved
  toward the storage box.}
  \caption{Cooperative task with the dual-arm ALOHA.}
  \label{fig:real-aloha}
\end{figure}

\begin{figure}[t]
  \centering
  \setlength{\abovecaptionskip}{3pt}
  \setlength{\belowcaptionskip}{0pt}
  \includegraphics[width=\linewidth]{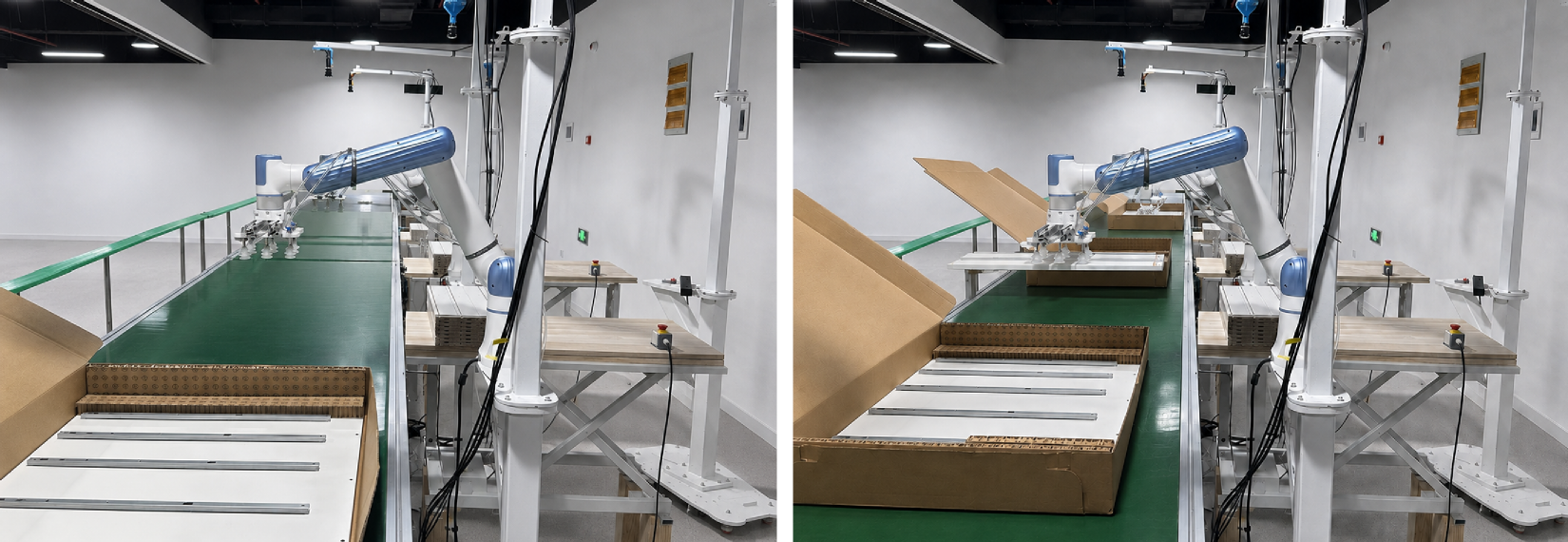}
  {\small (a) Conveyor setup and packing in progress.}
  \par\vspace{4pt}
  \includegraphics[width=\linewidth]{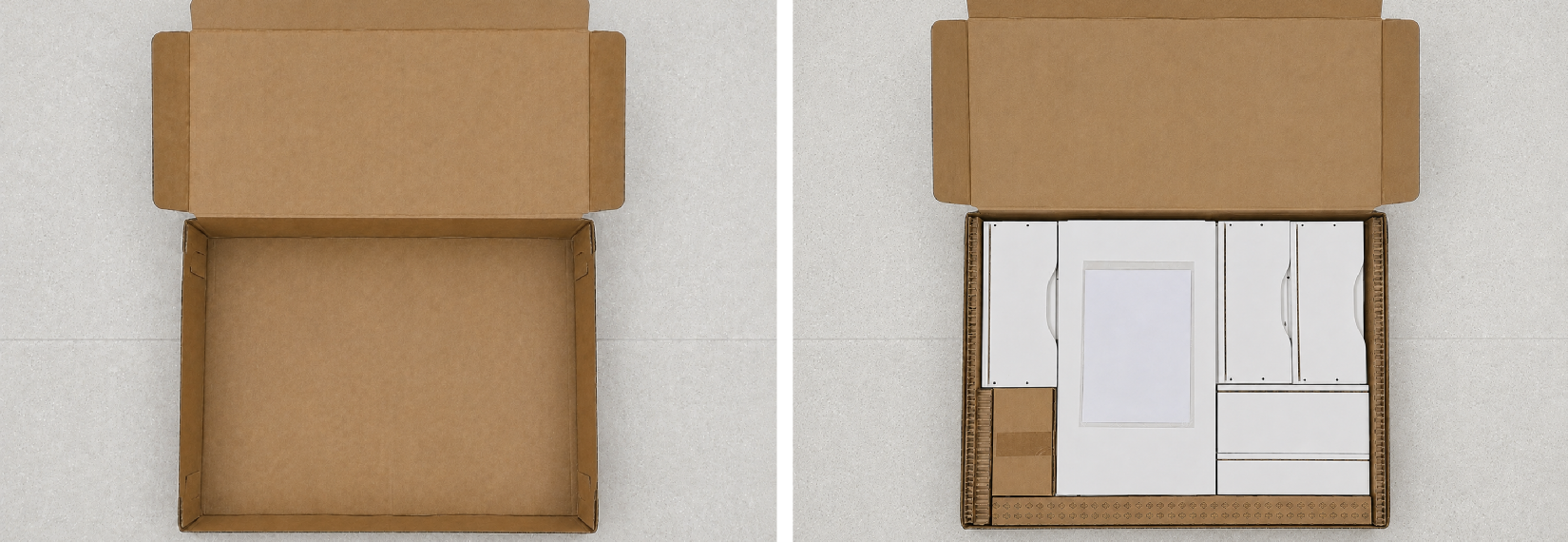}
  {\small (b) Box before and after packing.}
  \Description{The top panel shows the robot and moving conveyor during
  furniture packing. The bottom panel compares an empty box with a box
  containing aligned furniture boards and an instruction booklet.}
  \caption{Packing furniture components. The robot aligns
  boards with a moving box and adds a manual on top.}
  \label{fig:real-conveyor}
\end{figure}

\heading{Physical Tasks.}
Two physical tasks are included.
In cooperative pick-and-place, a dual-arm ALOHA~\cite{ALOHA} places two
beverage bottles into a box, varying lighting as well as bottle and box positions
across trials (Figure~\ref{fig:real-aloha}).
In assembly-line packing, Franka and DOBOT arms~\cite{fr3,dobot} pack
boards and instruction booklets for six furniture types into boxes on an
8\,m/min conveyor, with randomized lighting and box poses
(Figure~\ref{fig:real-conveyor}).
Both tasks run at 30\,Hz with 30-action chunks and 50 trials per method.

\heading{Baselines.}
torch.compile, vla.cpp~\cite{vlacpp}, and FlashRT~\cite{flashrt} are included as
inference-acceleration baselines that preserve model accuracy.
vla.cpp is unsupported on Jetson Thor and omitted there.
NIRVANA~\cite{nirvana}, DP-Cache~\cite{vlaperf2}, and \sys{} are all implemented on
FlashRT for $\pi_{0.5}$ and GR00T N1.6, and on torch.compile for X-VLA.
We also evaluate a variant that uses the previous inference as its
reference, following VLA-Cache~\cite{vla-cache}. We refer to this variant as \sys{} w/o Cache Repo.

\heading{Cache Settings.}
Table~\ref{tab:cache-pi05} shows the $\pi_{0.5}$--RoboTwin settings for the different methods.
DP-Cache uses default settings, with dense VLM prefill and the latest
denoising output reused at skipped steps.
Our NIRVANA adaptation keeps VLM prefill dense and reuses a prefix of reference
denoising outputs. It shares \sys{}'s retrieval method and reference
repository of about 8--32 episodes per task.
Profiling reduces NIRVANA's reused prefix to one step for GR00T N1.6 and
four for X-VLA, as half-schedule reuse degrades success.

\begin{table}[t]
  \centering\small
  \setlength{\abovecaptionskip}{5pt}
  \setlength{\belowcaptionskip}{0pt}
  \caption{Settings of different methods for $\pi_{0.5}$ on RoboTwin.}
  \label{tab:cache-pi05}
  \setlength{\tabcolsep}{3pt}
  \begin{tabular*}{\columnwidth}{@{\extracolsep{\fill}}llr@{}}
    \toprule
    Method & Setting & Value \\
    \midrule
    Shared & Denoising steps / chunk size & $10$ / $32$ \\
    DP-Cache & Cached steps & $\{0,1,2,6,8,9\}$ \\
    NIRVANA  & Reused steps & first $5$ \\
    \sys{} & $(\rho_{\mathrm{p}},\,\rho_{\mathrm{n}})$ & $(0.4,\,0.5)$ \\
    \bottomrule
  \end{tabular*}
\end{table}

\heading{Metrics.}
Latency spans from ready inputs to returned action chunks, including online retrieval
and weight gathering, after 20 warmups.
We report success rate (SR, \%) across all episodes.
Mean policy calls include only successful episodes, since failures
run to a step limit far beyond typical completion in both simulation
and physical experiments.
Vanilla denotes dense execution. Methods share initializations and checkpoints.

\begin{figure}[t]
  \centering
  \setlength{\abovecaptionskip}{3pt}
  \setlength{\belowcaptionskip}{0pt}
  \includegraphics[width=\columnwidth]{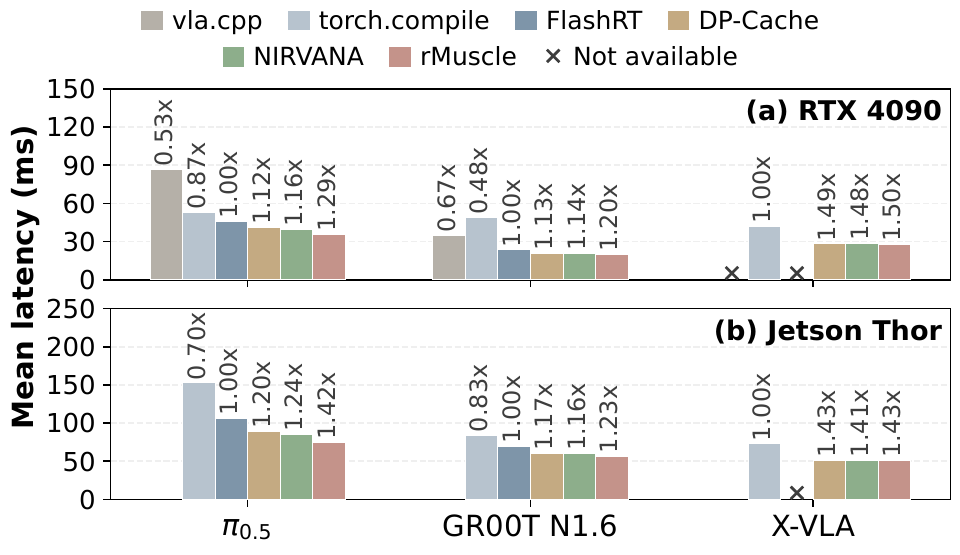}
  \Description{Two stacked panels of grouped bars show mean inference latency
  measured during benchmark evaluation on
  RTX 4090 (top) and Jetson Thor (bottom) for pi0.5, GR00T N1.6, and X-VLA,
  with speedups relative to FlashRT for pi0.5 and GR00T, and torch.compile
  for X-VLA. }
  \caption{Mean inference latency measured during benchmark
  evaluation on (a) RTX~4090 and (b) Jetson Thor.}
  \label{fig:hardware-performance}
\end{figure}

\subsection{Inference Performance}
\label{sec:eval-performance}

Figure~\ref{fig:hardware-performance} shows that \sys{} reduces latency
relative to each model's fastest exact baseline on both platforms.
Specifically, on RTX~4090, \sys{} reaches inference rates of
28.1, 51.0, and 35.7\,Hz for $\pi_{0.5}$, GR00T N1.6, and X-VLA, respectively. 
Speedups over the corresponding SOTA inference engine are $1.29\times$, $1.20\times$, and $1.50\times$.
On Jetson Thor, the corresponding inference rates are 13.4, 17.5, 
and 19.7\,Hz, with speedups of $1.42\times$, $1.23\times$, and $1.43\times$, respectively.

Compared with denoising step-skipping methods,
\sys{} also achieves inference speedup by accelerating both VLM prefill and action denoising.
Although step-skipping methods can achieve greater speedup in the denoising stage by aggressively skipping denoising steps,
this comes at the cost of reduced task success rates.

\heading{Scaling Model Sizes.} 
Scaling up VLA models can improve task performance~\cite{gen0}, but also
increases inference cost. We therefore examine how inference latency
and the speedup provided by \sys{} change with model size.
Keeping the visual tokenizer fixed, we scale both components within
the Gemma family: (VLM, denoise) sizes of (2B, 0.3B), (2B, 2B),
(7B, 2B), and (7B, 7B) correspond to $\pi_{0.5}$, $\pi_{0.5}$-L,
$\pi_{0.5}$-XL, and $\pi_{0.5}$-XXL, respectively.
For Gemma-7B, we use only the first 18 layers for visual-feature extraction.

\begin{figure}[t]
  \centering
  \setlength{\abovecaptionskip}{3pt}
  \setlength{\belowcaptionskip}{0pt}
  \includegraphics[width=\columnwidth]{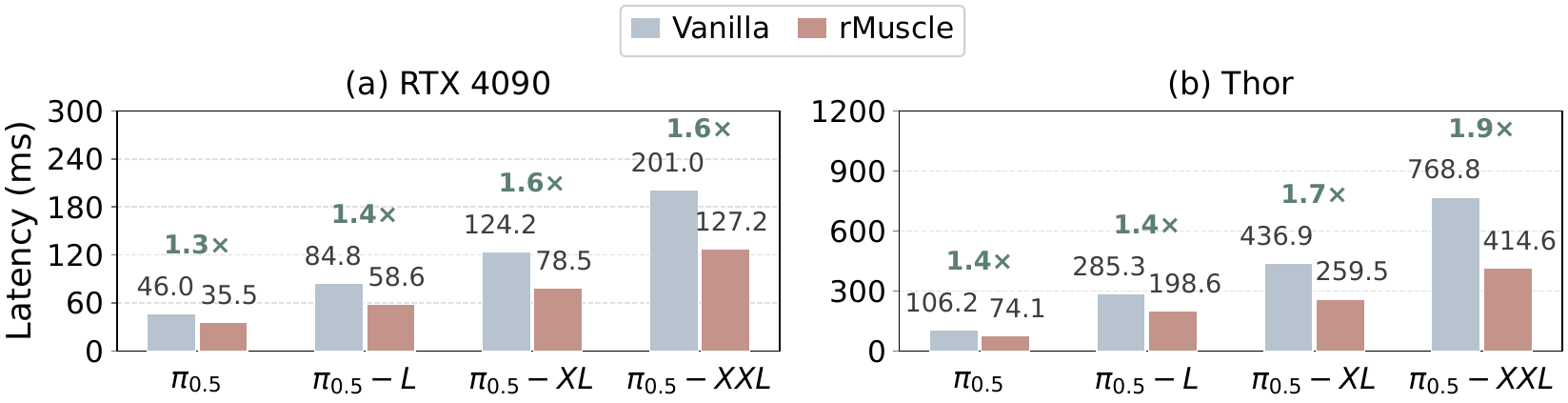}
  \Description{End-to-end latency for four VLM and denoiser size pairs
  on RTX 4090 and Jetson Thor. Speedups over the corresponding dense
  baseline are approximately 1.3, 1.4, 1.6, and 1.6 on RTX 4090,
  and 1.4, 1.4, 1.7, and 1.9 on Thor.}
  \caption{Inference latency of \sys{} across
  different model sizes on (a) RTX~4090 and (b) Jetson Thor.}
  \label{fig:model_scaling_speedup}
\end{figure}

Figure~\ref{fig:model_scaling_speedup} shows how speedup over the
FlashRT baseline changes with model size.
From the smallest to the largest configuration, speedup rises from
$1.3\times$ to $1.6\times$ on RTX~4090 and from $1.4\times$ to
$1.9\times$ on Thor. These results suggest that the dual-phase cache
can provide greater relative benefits as VLA models scale up.

\subsection{Policy Quality}
\label{sec:eval-quality}

\sys{} achieves the lowest latency among the compared methods.
We next evaluate its policy quality in simulation and on physical robots.

\begin{figure}[t]
  \centering
  \setlength{\abovecaptionskip}{3pt}
  \setlength{\belowcaptionskip}{0pt}
  \includegraphics[width=\columnwidth]{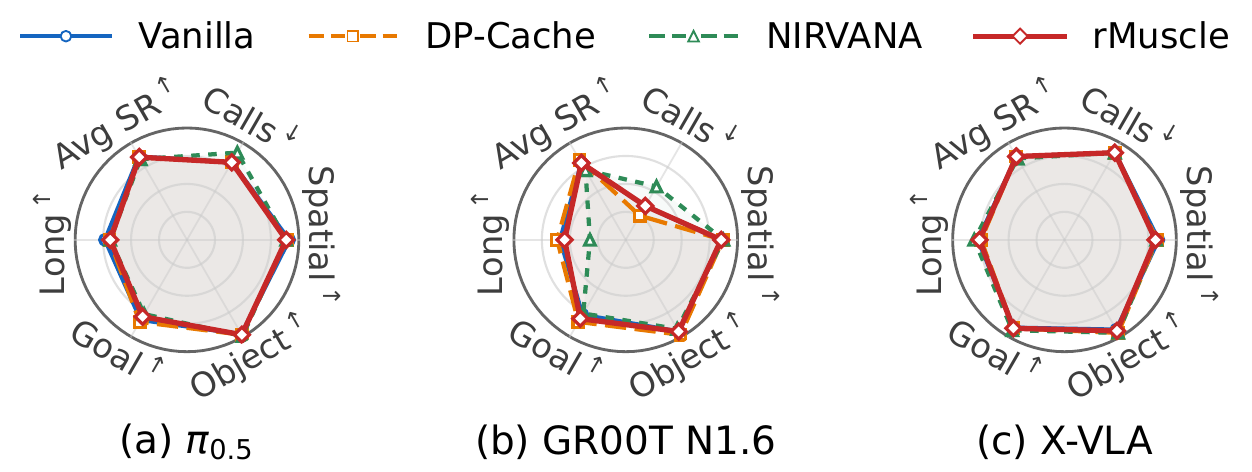}
  \caption{LIBERO results over four methods.
  SR axes span 85--100\%. Calls axes span 15--16, 21--23, and 5--6
  in panels (a)--(c), respectively.}
  \label{fig:libero-quality}
\end{figure}

\heading{Simulation.}
On LIBERO, Figure~\ref{fig:libero-quality} shows that \sys{} matches
the average success rates of vanilla and DP-Cache across
$\pi_{0.5}$, GR00T N1.6, and X-VLA.
It exceeds NIRVANA by 0.3, 1.1, and 0.3 points
on the three models, respectively.


\begin{table}[t]
  \centering\small
  \setlength{\abovecaptionskip}{5pt}
  \setlength{\belowcaptionskip}{0pt}
  \setlength{\tabcolsep}{3pt}
  \caption{RoboTwin results over all 50 tasks, with blue/pink shading for the best/worst values per row.}
  \label{tab:robotwin-quality}
  \resizebox{\columnwidth}{!}{%
  \begin{tabular}{@{}lcccc@{}}
    \toprule
    Task & Vanilla & DP-Cache & NIRVANA & \sys{} \\
    \midrule
    Click Alarmclock & 36 & \cellcolor{membound}34 & 35 & \cellcolor{cmpbound}42 \\
    Move Can Pot & \cellcolor{membound}22 & \cellcolor{membound}22 & 23 & \cellcolor{cmpbound}28 \\
    Adjust Bottle & \cellcolor{cmpbound}99 & 98 & \cellcolor{membound}86 & \cellcolor{cmpbound}99 \\
    Handover Block & \cellcolor{cmpbound}14 & \cellcolor{membound}4 & 9 & 12 \\
    Handover Mic & 45 & 31 & \cellcolor{membound}30 & \cellcolor{cmpbound}47 \\
    \multicolumn{5}{c}{\ldots\ (50 tasks)} \\
    Move Pillbottle Pad & 24 & \cellcolor{membound}20 & 21 & \cellcolor{cmpbound}28 \\
    Place Empty Cup & \cellcolor{cmpbound}39 & \cellcolor{cmpbound}39 & \cellcolor{membound}28 & 36 \\
    Dump Bin Bigbin & 58 & 59 & \cellcolor{membound}53 & \cellcolor{cmpbound}60 \\
    Stamp Seal & 17 & 17 & \cellcolor{membound}8 & \cellcolor{cmpbound}19 \\
    Turn Switch & 24 & 26 & \cellcolor{membound}20 & \cellcolor{cmpbound}28 \\
    \midrule
    Average SR (\%)\,\raisebox{0.15ex}{\scriptsize$\uparrow$} & \cellcolor{cmpbound}35.2 & 33.9 & \cellcolor{membound}26.9 & \cellcolor{cmpbound}35.2 \\
    Calls\,\raisebox{0.15ex}{\scriptsize$\downarrow$} & \cellcolor{cmpbound}7.0 & \cellcolor{membound}7.6 & 7.5 & 7.1 \\
    \bottomrule
  \end{tabular}}
\end{table}

\begin{table}[t]
  \centering\small
  \setlength{\abovecaptionskip}{5pt}
  \setlength{\belowcaptionskip}{0pt}
  \caption{$\pi_{0.5}$ on physical tasks. Blue/pink shading marks the best/worst values per row.}
  \label{tab:physical-quality}
  \setlength{\tabcolsep}{3pt}
  \begin{tabular*}{\columnwidth}{@{\extracolsep{\fill}}lcccc@{}}
    \toprule
    Metric & Vanilla & DP-Cache & NIRVANA & \sys{} \\
    \midrule
    \multicolumn{5}{c}{ALOHA: Cooperative bottle pick-and-place} \\
    \midrule
    SR (\%)\,\raisebox{0.15ex}{\scriptsize$\uparrow$} & \cellcolor{cmpbound}76 & 74 & \cellcolor{membound}70 & \cellcolor{cmpbound}76 \\
    Calls\,\raisebox{0.15ex}{\scriptsize$\downarrow$} & \cellcolor{cmpbound}38.7 & 39.5 & \cellcolor{membound}40.0 & 39.1 \\
    \midrule
    \multicolumn{5}{c}{DOBOT + FRANKA: Conveyor furniture packing} \\
    \midrule
    SR (\%)\,\raisebox{0.15ex}{\scriptsize$\uparrow$} & \cellcolor{cmpbound}84 & 80 & \cellcolor{membound}66 & \cellcolor{cmpbound}84 \\
    Calls\,\raisebox{0.15ex}{\scriptsize$\downarrow$} & \cellcolor{cmpbound}10.1 & 10.5 & \cellcolor{membound}10.7 & 10.2 \\
    \bottomrule
  \end{tabular*}
\end{table}

On RoboTwin, Table~\ref{tab:robotwin-quality} shows that \sys{} matches
vanilla's average success rate of 35.2\%, exceeding DP-Cache and
NIRVANA by 1.3 and 8.3 points, respectively.
It trails vanilla by 2 and 3 points on Handover Block and Place
Empty Cup due to corner cases with observation similarity near 0.8,
just above the fallback threshold.
In follow-up tests, raising this threshold or adding two relevant
reference episodes resolves these failures, while NIRVANA still fails.
These results highlight the importance of fallback and reference
coverage in preserving policy quality.

\heading{Physical Robots.}
Table~\ref{tab:physical-quality} shows that \sys{} matches vanilla
execution on both physical tasks: 76\% on ALOHA cooperative bottle
pick-and-place and 84\% on DOBOT and Franka conveyor packing.
It exceeds DP-Cache by 2 and 4 points, and NIRVANA by 6 and 18
points on the two tasks, respectively.

\subsection{Memory Overhead}
\label{sec:eval-memory}

We quantify the CPU cache footprint of \sys{} and compare its GPU
memory use with vanilla execution.
To represent increasing reference requirements for more difficult tasks,
we increase the per-task cache capacity from 200 entries across
8 reference episodes to 800 entries across 32 reference episodes,
while expanding the candidate window from $T=L=1$ to $T=L=4$.

\begin{figure}[t]
  \centering
  \setlength{\abovecaptionskip}{3pt}
  \setlength{\belowcaptionskip}{0pt}
  \includegraphics[width=\columnwidth]{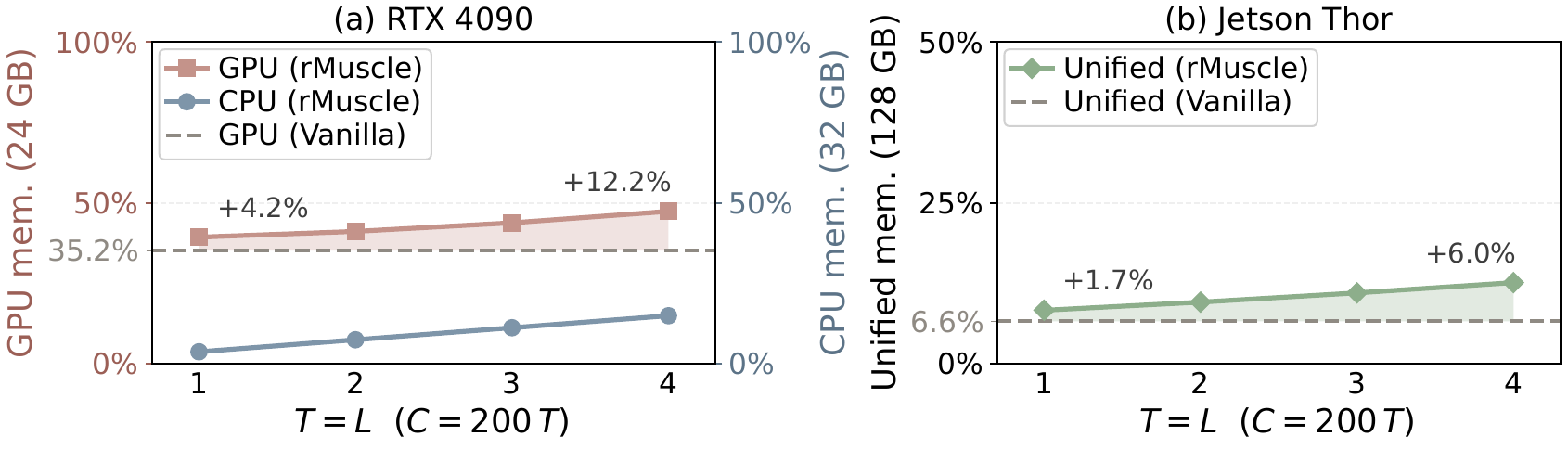}
  \Description{Memory use as a share of device memory across four cache
  configurations. On RTX 4090, GPU memory rises from 39.4\% to 47.4\% of
  24 GB, compared with 35.2\% for vanilla, an increase of 4.2 to 12.2
  percentage points, while the CPU cache occupies 3.7\% to 14.9\% of
  32 GB. On Jetson Thor, unified memory rises from 8.3\% to 12.6\% of
  128 GB, compared with 6.6\% for vanilla, an increase of 1.7 to 6.0
  percentage points.}
  \caption{Memory use as a share of device memory on (a) RTX~4090 and
  (b) Jetson Thor.
  $T$ and $L$ determine the GPU candidate working-set size, $T(L+1)$.
  $C$ denotes the total number of cached inference entries for
  the current task.}
  \label{fig:memory-overhead}
\end{figure}

Figure~\ref{fig:memory-overhead} shows that on RTX~4090,
the CPU cache accounts for 3.7--14.9\% of 32\,GB host memory.
GPU memory increases from 8.4\,GB for vanilla to 9.4--11.4\,GB,
raising occupancy by less than 13\%.
On Jetson Thor, where CPU and GPU share 128\,GB unified memory,
occupancy increases from 6.6\% to 8.3--12.6\%.
Overall, \sys{} supports up to 800 cached inference entries with less than 3\,GB additional GPU memory on RTX~4090
and at most 6.0 percentage points additional unified-memory occupancy on Jetson Thor.

\subsection{Ablation Studies}
\label{sec:eval-ablation}

We evaluate cache contributions and quality--latency tradeoffs using $\pi_{0.5}$ on RoboTwin.

\begin{figure}[t]
  \centering
  \setlength{\abovecaptionskip}{3pt}
  \setlength{\belowcaptionskip}{0pt}
  \includegraphics[width=\columnwidth]{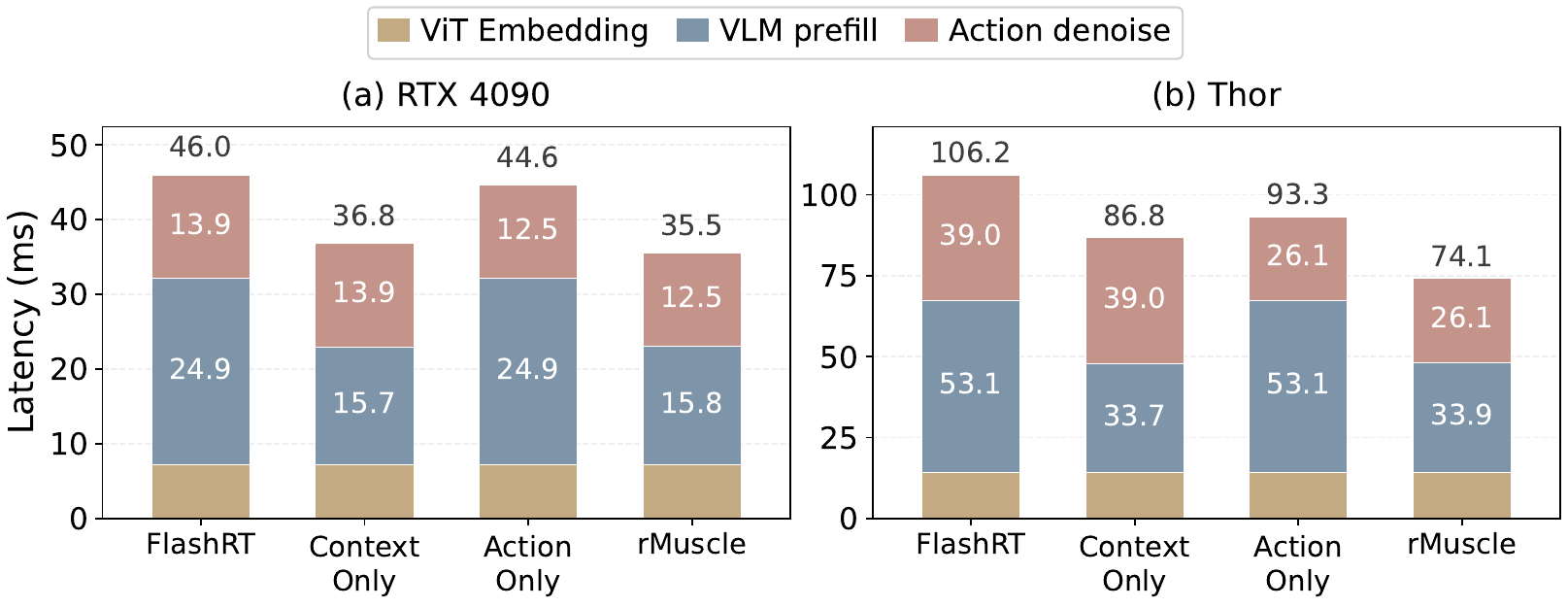}
  \caption{Latency breakdown for VLA inference phases.}
  \label{fig:component-ablation}
\end{figure}

\heading{Cache Contributions.}
We compare FlashRT, Context Cache alone, Action Cache alone, and
the full system in Figure~\ref{fig:component-ablation} to isolate the
latency contribution of each cache on RTX~4090 and Jetson Thor. 


Context Cache alone gives the larger end-to-end
speedup: $1.25\times$ on RTX~4090 and $1.22\times$ on Jetson Thor versus FlashRT.
With Context Cache, the reduction in VLM prefill latency accounts for 87\% of the total latency reduction on RTX~4090, compared with 60\% on Jetson Thor.
Action Cache alone speeds up action denoising by $1.11\times$ on RTX~4090
and $1.49\times$ on Jetson Thor, consistent with the more memory-bound
Thor profile in Table~\ref{tab:roofline}.

\begin{figure}[t]
  \centering
  \setlength{\abovecaptionskip}{3pt}
  \setlength{\belowcaptionskip}{0pt}
  \includegraphics[width=\columnwidth]{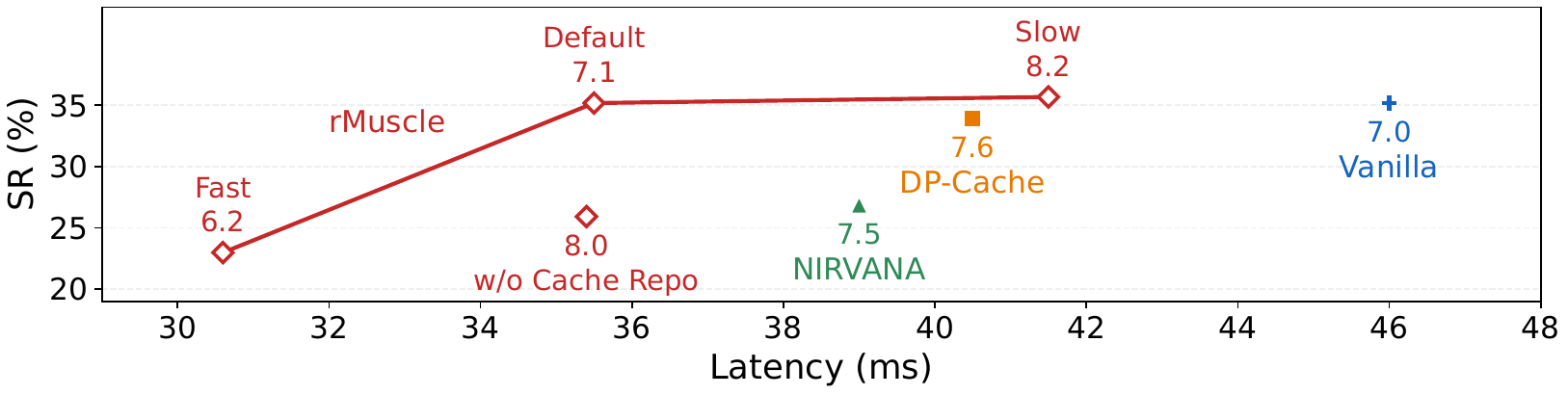}
  \caption{Quality--latency tradeoff on RTX~4090.
  Numeric labels report average calls.
  $(\rho_{\mathrm{p}},\rho_{\mathrm{n}})$ configurations:
  Fast $(0.2,0.25)$, Default $(0.4,0.50)$, Slow $(0.6,0.75)$.}
  \label{fig:tradeoff-sweep}
\end{figure}

\heading{Reference Repository.}
The \sys{} w/o Cache Repo variant uses the previous policy call
within the same episode as its reference instead of retrieving one
from the repository.
As shown in Figure~\ref{fig:tradeoff-sweep}, Default achieves
a success rate of 35.2\%, compared with 25.9\% without the
reference repository, at similar latency.
These results highlight the value of references from prior executions
in preserving policy quality, while the similar latency suggests that
cache management adds negligible overhead to the critical path.

\heading{Quality--Performance Tradeoff.}
\label{sec:eval-tradeoff}
We vary the visual-token recomputation ratio and the neuron fraction of \sys{}.
Figure~\ref{fig:tradeoff-sweep} compares the Fast, Default, and Slow
configurations on RTX~4090 alongside Vanilla, DP-Cache, NIRVANA,
and the repository ablation discussed above.

Default increases success from 23.0\% to 35.2\% compared with Fast,
at an additional latency of 4.9\,ms.
Default matches Vanilla's success rate, while Slow achieves a slightly higher success rate.
They reduce latency relative to Vanilla by 10.5\,ms and 4.5\,ms, respectively.
These results meet our design goal of preserving policy quality while
reducing inference latency.
Consistent with our profiling results, Default offers the lowest
latency without sacrificing success.

\section{Related Work}


\noindentparagraph{VLA Profiling and Inference Engines.}
VLA-Perf analytically explores how architecture, hardware placement, and
serving choices affect inference latency~\cite{vlaperf}, while a cross-XPU
study measures energy and latency across heterogeneous accelerators and
validates DP-Cache on LIBERO and a physical Franka robot~\cite{vlaperf2}.

Inference engines optimize how model operations execute on the
target hardware. The vla.cpp runtime provides a portable C++ implementation for multiple VLA
architectures and hardware tiers, including native support for iterative
diffusion and flow-matching action heads~\cite{vlacpp}.  RealVLA removes
inference overhead and integrates the optimized model path with streaming robot
control~\cite{realvla}.  FlashRT uses specialized kernels, operator fusion, and
whole-forward CUDA Graph replay, together with precision-specific execution
modes, for latency-sensitive VLA serving~\cite{flashrt}.  These systems reduce
framework, kernel, and scheduling overhead without determining which visual
tokens or denoising computations may be reused.

\noindentparagraph{Approaches to VLA Acceleration.}
VLA accelerators reduce multimodal or action-generation work through pruning,
approximate speculative execution, and efficient model design~\cite{Efficientsurvey}.  
VLA-Cache accelerates attention computation by reusing cached KV states
of static visual tokens from the preceding frame~\cite{vla-cache}.
However, attention-only reuse does not target the dominant FFN bottlenecks
in the state-of-the-art VLA models studied here.

SpecPrune-VLA uses recent history and action-aware control to prune visual tokens, 
while Realtime-VLA FLASH combines preceding visual context 
with draft-model speculation and action-expert verification~\cite{SpecPrune,flashvla}.
Neither searches a persistent repository.  
For action generation, TS-DP couples a distilled diffusion drafter with a learned scheduler~\cite{TS-DP}, 
whereas RoboMamba builds an efficient policy around a Mamba state-space backbone and a
lightweight policy head~\cite{RoboMamba}.  

\section{Discussion}
\label{sec:discussion}

\noindentparagraph{Why Inference Latency Matters.}
Policy latency is the gap between the last action of one chunk and the first action of the next,
while actions within a chunk are executed at 20--30 Hz.
As policy-call frequency approaches the execution rate,
chunk boundaries become smoother, and the robot can re-observe and react to environmental changes more frequently.
Lower inference latency therefore improves responsiveness, motion smoothness, and task throughput.
Since each episode requires 4--40 policy calls,
reducing per-call latency shortens the overall task cycle and
enables a fixed-workstation robot to complete more tasks per hour on the same hardware.

\noindentparagraph{Dynamic Cache Update.}
Unseen tasks initially run densely, and successful episodes naturally
provide new cache references at no additional computation cost.
Successful episodes from underrepresented observation clusters could be admitted.
A new task variant therefore transitions from dense execution to
cached acceleration as its reference coverage improves.
We leave online cache updates for future work.

\section{Conclusion}

This paper introduces \sys{}, the fastest VLA inference engine optimized for embodied workloads.
Through an in-depth analysis of robot execution patterns,
we find that repeated task executions exhibit similarity not only in observations and action trajectories,
but also in internal model states throughout inference.
\sys{} exploits these similarities through a dual-cache design that
reduces VLM computation and denoising weight traffic.
It further combines online cache reconstruction, sliding-window cache retrieval,
and mask sharing across denoising steps to
keep the memory footprint and cache-access overhead negligible.
\sys{} achieves a $1.29$--$1.42\times$ end-to-end speedup on RTX~4090 and Jetson Thor
over the state-of-the-art inference engine,
while preserving the original success rates on both simulation benchmarks and physical-robot tasks.

\bibliographystyle{ACM-Reference-Format}
\bibliography{refs}

\end{document}